%% file: main.tex
\documentclass[lettersize,journal]{IEEEtran}
\ifCLASSOPTIONcompsoc
  \usepackage[nocompress]{cite}
\else
  \usepackage{cite}
\fi
\ifCLASSINFOpdf
\else
\fi
\usepackage{subfigure}
\usepackage{algorithm}
\usepackage{algorithmic}
\usepackage{xcolor}
\usepackage{enumitem}
\usepackage{amsmath}
\usepackage{bbm}
\usepackage{amsfonts}
\usepackage{graphicx}

\newcommand{{\method}}{GraphSMOTE}
\DeclareMathOperator*{\argmin}{argmin}
\DeclareMathOperator*{\argmax}{argmax}

\begin{document}
\title{Synthetic Over-sampling for Imbalanced Node Classification with Graph Neural Networks }

\author{Tianxiang Zhao,
        Xiang Zhang,
        and~Suhang Wang% <-this % stops a space
\thanks{ T. Zhao, X. Zhang and S. Wang are with the College of Information Sciences and Technology, The Pennsylvania State University, University Park, PA 16802.\protect\\
E-mail: \{tkz5084, xzz89, szw494\}@psu.edu}% <-this % stops an unwanted space
\thanks{Manuscript received April 19, 2005; revised August 26, 2015.}}


\markboth{Journal of \LaTeX\ Class Files,~Vol.~14, No.~8, August~2015}%
{Shell \MakeLowercase{\textit{et al.}}: A Sample Article Using IEEEtran.cls for IEEE Journals}
\maketitle

\begin{abstract}
In recent years, graph neural networks (GNNs) have achieved state-of-the-art performance for node classification. However, most existing GNNs would suffer from the graph imbalance problem. In many real-world scenarios, node classes are imbalanced, with some majority classes making up most parts of the graph. The message propagation mechanism in GNNs would further amplify the dominance of those majority classes, resulting in sub-optimal classification performance. In this work, we seek to address this problem by generating pseudo instances of minority classes to balance the training data, extending previous over-sampling-based techniques. This task is non-trivial, as those techniques are designed with the assumption that instances are independent. Neglection of relation information would complicate this oversampling process. Furthermore, the node classification task typically takes the semi-supervised setting with only a few labeled nodes, providing insufficient supervision for the generation of minority instances. Generated new nodes of low quality would harm the trained classifier. In this work, we address these difficulties by synthesizing new nodes in a constructed embedding space, which encodes both node attributes and topology information. Furthermore, an edge generator is trained simultaneously to model the graph structure and provide relations for new samples. To further improve the data efficiency, we also explore synthesizing mixed ``in-between'' nodes to utilize nodes from the majority class in this over-sampling process. Experiments on real-world datasets validate the effectiveness of our proposed framework.
\end{abstract}


\section{Introduction}
\IEEEPARstart{R}{ecent} years have witnessed great improvements in learning from graphs with the developments of graph neural networks (GNNs)~\cite{Kipf2017SemiSupervisedCW,Hamilton2017InductiveRL,xu2018powerful}. One typical research topic is semi-supervised node classification~\cite{yang2016revisiting}, in which we have a large graph with a small ratio of nodes labeled. This task requires training a classifier on those supervised nodes, and then use it to predict labels of other nodes during testing. GNNs have obtained state-of-the-art performance in this task, and is developing rapidly. For example, GCN~\cite{Kipf2017SemiSupervisedCW} exploits features in the spectral domain efficiently by using a simplified first-order approximation; GraphSage~\cite{Hamilton2017InductiveRL} utilizes features in the spatial domain and is better at adapting to diverse graph topology. %GIN~\cite{xu2018powerful} extends the representation ability of GNN to that of WL test; etc.  
Despite all the progress, existing works mainly focus on the balanced setting, with different node classes of similar sizes.

\begin{figure}[t!]
  \centering
  \subfigure[Bot detection task]{
		\label{fig:example-bot}
		\includegraphics[width=0.22\textwidth]{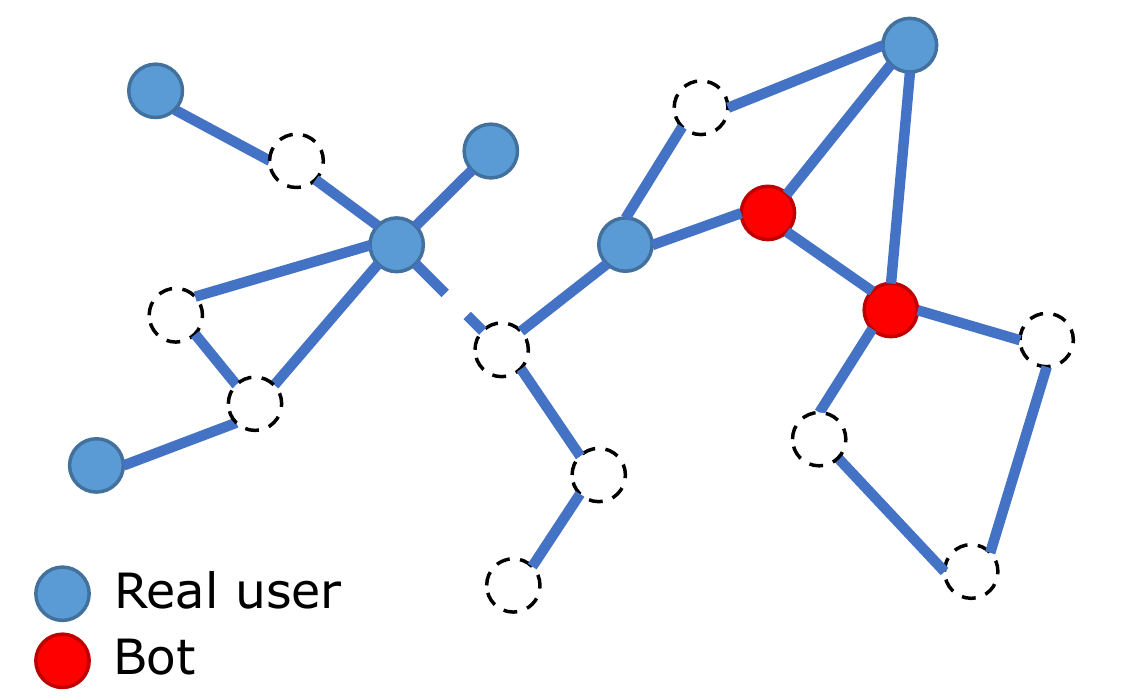}}
    \subfigure[After over-sampling]{
		\label{fig:example-after}
		\includegraphics[width=0.22\textwidth]{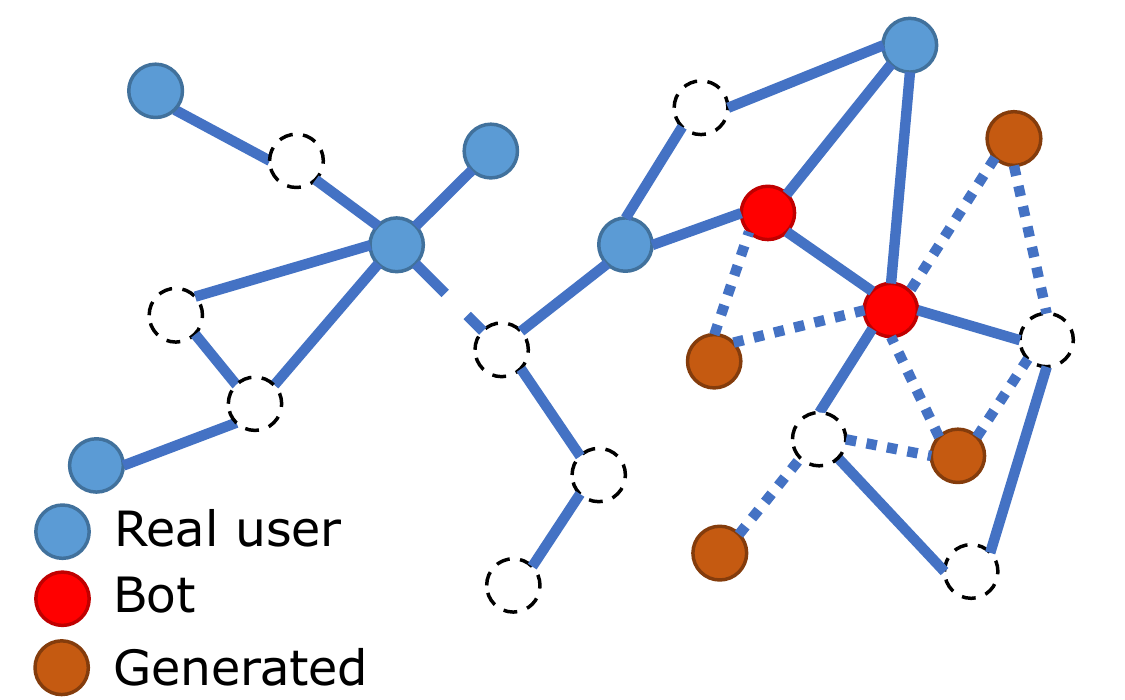}}
		
    \vskip -1em
    \caption{An example of bot detection on a social network, and the idea of over-sampling. Blue nodes are real users, red nodes are bots, and nodes in dots are unlabeled. Through over-sampling, pseudo nodes of minority class are generated(shown in color brown) to make node classes balanced. Note that the over-sampling is in the latent space.} \label{fig:example}
  \setlength{\abovecaptionskip}{0cm}
  \vskip -1em
\end{figure}

In many real-world applications, node classes could be imbalanced in graphs, i.e., some classes have significantly fewer samples for training than other classes. For example, for fake account detection~\cite{mohammadrezaei2018identifying,zhao2009botgraph}, the majority of users in a social network platform are benign users while only a small portion of them are bots. Similarly, topic classification for website pages~\cite{wang2020network} could also suffer from this problem, as the materials for some topics are scarce, compared to those on-trend topics. %Other examples include malicious machine detection~\cite{Stringhini2015EVILCOHORTDC}, recommendation effects analysis in social communities~\cite{stoica2019hegemony}, etc. 
Thus, we are often faced with the imbalanced node classification problem. An example of the imbalanced node classification problem is shown in Figure~\ref{fig:example-bot}.  Each blue node refers to a real user, each red node refers to a fake user, and the edges denote the friendship. The task is to predict whether those unlabeled users in dashes are real or fake. The classes are imbalanced in nature, as fake users often make up a small ratio of all the users~\cite{gurajala2015fake}. The semi-supervised setting further magnifies the class imbalanced issue as we are only given limited labeled data, which makes the number of labeled minority samples extremely small. The imbalanced node classification brings challenges to existing GNNs because the majority classes could dominate the loss function of GNNs, which makes the trained GNNs over-classify those majority classes and become unable to predict accurately for samples from minority classes. This issue impedes the adoption of GNNs for many real-world applications with imbalanced class distribution such as malicious account detection. Therefore, it is important to develop GNNs for class imbalanced node classification.

In machine learning domain, traditional class imbalance problem has been extensively studied. Existing algorithms can be summarized into three groups: data-level approaches, algorithm-level approaches, and hybrid approaches. Data-level approaches seek to make the class distribution more balanced, using over-sampling or down-sampling techniques\cite{more2016survey,chawla2002smote}; algorithm-level approaches typically introduce different mis-classification penalties or prior probabilities for different classes~\cite{elkan2001foundations,ling2008cost,zhou2005training}; and hybrid approaches~\cite{chawla2003smoteboost,liu2008exploratory} combine both of them. Among these algorithms, data-level approaches, especially over-sampling-based like Synthetic Minority Oversampling TEchnique (SMOTE)~\cite{chawla2002smote}, have shown to perform both effectively and stably~\cite{buda2018systematic,johnson2019survey}. SMOTE augments training data through generating new instances of minority classes via interpolation. However, it is designed from the independent and identical distribution (i.i.d) assumption and is unsuitable for relational data structures.
%directly applying existing algorithms to graphs may get sub-optimal results, due to their independent and identical distribution (i.i.d) assumption of taking each sample as independent. 
There are some pioneering works for imbalanced node classification~\cite{chen2021topology,wang2021distance,shi2020multi,wu2021graphmixup}, but they mainly rely on re-weighting or few-shot learning and are essentially different from our objective: extending SMOTE to augment imbalanced graphs.

In this work, we propose to extend existing over-sampling techniques like SMOTE to imbalanced node classification with GNNs~\footnote{Code available at https://github.com/TianxiangZhao/GraphSmote}. The idea is shown in Figure~\ref{fig:example-after}. Traditional imbalanced learning algorithms are not readily applicable to graphs, mainly due to two-folded reasons. \textit{First}, it is difficult to generate relation information for synthesized new samples. Mainstream over-sampling techniques~\cite{more2016survey} use interpolation between target example and its nearest neighbor to generate new training examples. However, interpolation is improper for edges, as they are usually discreet and sparse. Interpolation could break down the topology structure. \textit{Second}, synthesized new samples could be of low quality. Node attributes are high-dimensional, and topology information need to be encoded for measuring similarity. Directly interpolating on node attributes would easily generate out-of-domain examples, which are not beneficial for training the classifier. 

Targeting at aforementioned problems, we extend previous over-sampling algorithms to a new framework in order to cope with graphs. Concretely, we take SMOTE as the base approach, and annotate the new framework as {\method}. The modifications are mainly at two places. First, we propose to obtain new edges between generated samples and existing samples with an edge predictor. This predictor can learn the genuine distribution of edges, and hence can be used to produce reliable relation information among samples. Second, we propose to perform interpolation at the intermediate embedding space learned by a GNN network so that both node attributes and local topology information are encoded, inspired by ~\cite{ando2017deep}. In this intermediate embedding space, the dimensionality is much lower, and the distribution of samples from the same class would be more dense. As intra-class similarity as well as inter-class differences would have been captured, interpolation can be better trusted to generate in-domain samples. Concretely, we propose a new framework in which graph auto-encoding task and node classification task are combined together. These two tasks share the same feature extractor, and over-sampling is performed at the output space of that module, as shown in Figure~\ref{fig:model_architecture}. 

Besides, another difficulty that stands out in semi-supervised node classification is the lacking of training data. For example, in the bot detection example from Figure~\ref{fig:example-bot}, only a small ratio of nodes will be labeled for training. This fact would further amplify the difficulty of learning from minority classes. Vanilla SMOTE strategy~\cite{chawla2002smote} only utilizes data from the minority classes, while leaving the vast number of nodes from the majority classes untouched. As nodes from the minority classes are limited, great potential could lie behind incorporating knowledge from those majority nodes in the data augmentation process. Targeting at this problem, we further explore the utilization of majority nodes in graph augmentation. Concretely, we generate ``in-between'' nodes through performing interpolation also on node pairs from different classes, inspired by Mixup~\cite{zhang2017mixup}. Mixup is a commonly-used regularization technique, which trains the model with mixed instances generated through convex combinations in both feature and label spaces. It has been found to be capable of making different classes more disparate~\cite{chou2020remix}. We extend this technique to the graph domain to generate mixed nodes, supervise these generated mixed nodes with mixed pseudo labels, and also use them to augment the training data. This extension will further augment {\method}, and provides the classifier with more signals on the class boundary. The main contributions of the paper are:
\begin{itemize}
    \item We study a novel problem of imbalanced node classification by generating synthetic nodes for minority classes to augment the graph. 
    \item We design a new framework {\method} which extends over-sampling algorithms to work for graph data. It addresses the deficiencies of previous methods, by generating more natural nodes as well as relation information. Besides, it is general and easy to extend.
    \item We further extend {\method} to augment the training data with not only nodes from minority classes, but also those from majority classes with the mixup technique. It is important in few-label scenarios as minority classes may have very few instances and effective augmentation based only on them would be difficult.
    \item Experiments results show that {\method} outperforms all baselines with a large gap. The effectiveness of extended mixup is also evaluated in the few-label case. Extensive analysis of our model's behavior as well as recommended settings are also presented.
\end{itemize}

\input{related_work.tex}

\input{problem_setting.tex}

\input{methodology.tex}

\input{experiment.tex}

\section{Conclusion and Future Work} \label{sec:conclusion}

Class imbalance problem of nodes in graphs widely exists in real-world tasks, like fake user detection, web page classification, malicious machine detection, etc. This problem can significantly influence classifier's performance on those minority classes, but is left unconsidered in previous works. Thus, in this work, we investigate this imbalanced node classification task. Specifically, we propose a novel framework {\method}, which extends previous over-sampling algorithms for i.i.d data to this graph setting. Concretely, {\method} constructs an intermediate embedding space with a feature extractor, and train an edge generator and a GNN-based node classifier simultaneously on top of that. Experiments on one artificial dataset and two real-world datasets demonstrated its effectiveness, outperforming all other baselines with a large margin. Ablation studies are performed to understand {\method} performs under various scenarios. Parameter sensitivity analysis is also conducted to understand the sensitivity of {\method} on the hyperparameters. To cope with semi-supervision setting and the lack of minority nodes, we further consider the utilization of majority nodes in augmenting the graph. An extension is made on {\method} to generate mixed nodes, through interpolating node pairs from different classes. Experiments are conducted to evaluate the advantage of this augmentation technique, and it is shown to improve more in the few-label cases or more imbalanced settings.

There are several interesting directions need further investigation. First, besides node classification, other tasks like edge type prediction or node representation learning may also suffer from under-representation of nodes in minority classes. And sometimes, node class might not be provided explicitly. Therefore, we will also extend {\method} for handling other types of imbalanced learning problems on graphs. Second, in this paper, we mainly conduct experiments on citation network and social media network. There are many other real-world applications which can be treated as imbalanced node classification problems. Therefore, we would like to extend our framework for more application domains such as document analysis in the websites.

% or


%\appendices

%\section{}

\ifCLASSOPTIONcompsoc
  \section*{Acknowledgments}
\else
  \section*{Acknowledgment}
\fi

This material is based upon work supported by, or in part by, the National Science Foundation under grants number IIS-1707548, CBET-1638320, IIS-1909702, IIS1955851, and the Global Research Outreach program of Samsung Advanced Institute of Technology under grant \#225003. The findings and conclusions in this paper do not necessarily reflect the view of the funding agency.

\ifCLASSOPTIONcaptionsoff
  \newpage
\fi



%\bibliographystyle{IEEEtran}

%\begin{thebibliography}{1}


%\end{thebibliography}
\bibliographystyle{IEEEtran}
\bibliography{bib1}


\begin{IEEEbiography}[{\includegraphics[width=1in,height=1.25in,clip,keepaspectratio]{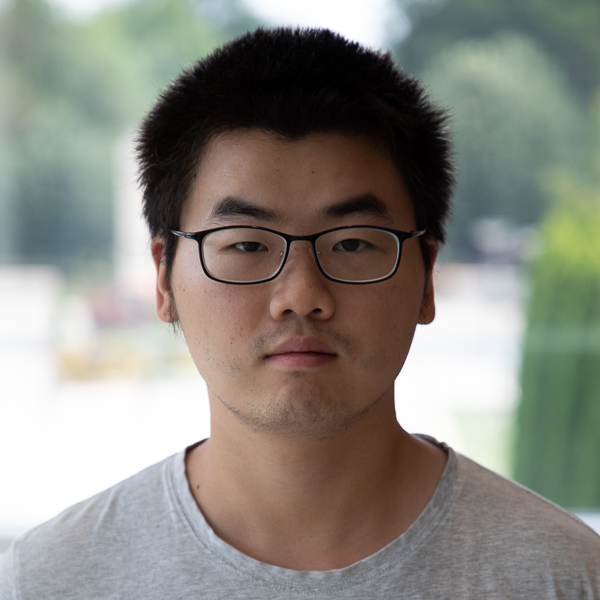}}]{Tianxiang Zhao}
received his BS degree in computer science from University of Science and Technology of China (USTC), Hefei, China,
in 2017. He
is currently working toward the Ph.D. degree in
IST at the Pennsylvania State University (PSU), State College,
PA, under the supervision of d Dr. Suhang Wang and Dr. Xiang Zhang since 2019.
His research interests are in graph neural networks, weak supervision tasks and knowledge transfer. He also worked as a
research intern at NEC in 2021.
\end{IEEEbiography}

\begin{IEEEbiography}[{\includegraphics[width=1in,height=1.25in,clip,keepaspectratio]{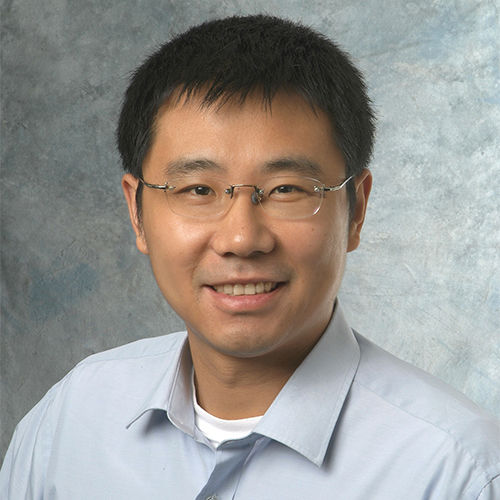}}]{Xiang Zhang}
is an Associate Professor in the College of Information Sciences and Technology at the Pennsylvania State University. His research bridges the areas of big data, data mining, machine learning, database and biomedical informatics. He is particularly interested in developing algorithms and models for analyzing large data sets generated in social, biological and medical domains.
\end{IEEEbiography}

%\newpage

\begin{IEEEbiography}[{\includegraphics[width=1in,height=1.25in,clip,keepaspectratio]{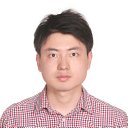}}]{Suhang Wang}
is an assistant professor of the College of Information Sciences and Technology at the Pennsylvania State University. He received his Ph.D. in Computer Science from Arizona State University in 2018, M.S. in Electrical Engineering from University of Michigan - Ann Arbor in 2013, and B.S. in Electrical and Computer Engineering from Shanghai Jiao Tong University in 2012. His research interests are in graph mining, data mining and machine learning. He is an associate editor for several journals and serves as regular journal reviewers and numerous conference program committees. He has published innovative works in highly ranked journals and top conference
proceedings such as IEEE TKDE, ACM TIST, KDD, WWW, AAAI, IJCAI, CIKM, SDM, WSDM, ICDM and CVPR, which have received extensive coverage in the media.
\end{IEEEbiography}


%\vfill


\end{document}

%% file: related_work.tex
\section{Related Work} \label{sec:related_work}
In this section, we briefly review related works, which include graph neural networks and class imbalance problem.

\subsection{Class Imbalance Problem}
Class imbalance is common in real-world applications, and has long been a classical research direction in the machine learning domain. Plenty of tasks suffer from this problem, like medical diagnosis~\cite{mac2002problem,grzymala2004approach} or fine-grained image classification~\cite{peng2017object,van2017inaturalist}. Classes with larger number of instances are usually called as majority classes, and those with fewer instances are usually called as minority classes. The countermeasures against this problem can generally be classified into three groups, i.e., algorithm-level, data-level and hybrid. %Algorithm level approaches generally address the problem class imbala
Data-level approaches seek to directly adjust class sizes through over- or under-sampling ~\cite{kubat1997addressing,barandela2004imbalanced}. For example, the vanilla form of over-sampling is simply replicate minority data samples to balance the dataset. However, it might lead to over-fitting as no extra information is introduced. SMOTE~\cite{chawla2002smote} addresses this problem by generating new samples, performing interpolation between samples in minority classes and their nearest neighbors. SMOTE is the most popular over-sampling approach, and many extensions are proposed on top of it to make the interpolation process more effective, such as Borderline-SMOTE~\cite{han2005borderline}, EmbSMOTE~\cite{ando2017deep} and Safe-Level-SMOTE~\cite{bunkhumpornpat2009safe}. %Cluster-based Over-sampling~\cite{jo2004class} first clusters samples into different groups, then over-samples each group separately, considering that small districts often exist in the input space. 
Algorithm-level approaches adopt class-specific losses or incorporate distribution priors~\cite{ling2008cost,lawrence1998neural}. For example, cost sensitive learning~\cite{zhou2005training,ling2008cost} generally constructs a cost matrix to assign different mis-classification penalties for different classes. ~\cite{parambath2014optimizing} proposes an approximation to F measurement, which can be directly optimized by gradient propagation.  Threshold moving~\cite{lawrence1998neural} modifies the inference process after the classifier is trained, by introducing a prior probability for each class.  Hybrid approaches~\cite{liu2008exploratory,chawla2003smoteboost,he2009learning,sun2007cost} combine multiple algorithms from one or both aforementioned categories. %~\cite{liu2008exploratory} uses a group of classifiers, each one is trained on a subset of majority classes and minority classes. ~\cite{chawla2003smoteboost} combines boosting with SMOTE approach, and ~\cite{he2009learning} combines over-sampling with cost sensitive learning. ~\cite{sun2007cost} introduces three cost-sensitive boosting approaches, which iteratively updates the impact of each class in together with the AdaBoost parameters.

%The aforementioned approaches are designed for i.i.d data; while the work on imbalanced node classification on graphs are less explored. Recently, some efforts have been made to improve the imbalanced node classification~\cite{wu2021graphmixup, chen2021topology, shi2020multi, wang2021distance,wang2020network}. For instance, DPGNN~\cite{wang2021distance} proposes a class prototype-driven training loss to maintain the balance of different classes. ReNode~\cite{chen2021topology} re-weights each training node via examining the propagated label information to their neighbors.

%Our work are inherently different from existing work. First, synthetic minority oversampling techniques such as SMOTE are popular and effective approaches for addressing  class imbalance on i.i.d data ~\cite{buda2018systematic,johnson2019survey}. However, they cannot be directly applied to graph structured data because: (i) the synthetic node generation on the raw feature space cannot take the graph information into consideration; and (ii) the generated nodes doesn't have links with the graph, which cannot facilitate the graph based classifier such as GNNs. Hence, in this work, we focus on extending SMOTE into graph domain for GNNs.
Some systematic analysis found that synthetic minority oversampling techniques such as SMOTE are popular and effective approaches for addressing  class imbalance~\cite{buda2018systematic,johnson2019survey}. However, existing work are overwhelmingly dedicated to i.i.d data. They cannot be directly applied to graph structured data because: (i) the synthetic node generation on the raw feature space cannot take the graph information into consideration; and (ii) the generated nodes doesn't have links with the graph, which cannot facilitate the graph based classifier such as GNNs. Hence, in this work, we focus on extending SMOTE into graph domain for GNNs. 

Recently, some efforts have been made to improve the imbalanced node classification~\cite{wu2021graphmixup, chen2021topology, shi2020multi, wang2021distance,wang2020network}. For instance, DPGNN~\cite{wang2021distance} proposes a class prototype-driven training loss to maintain the balance of different classes. ReNode~\cite{chen2021topology} re-weights each training node via examining the propagated label information to their neighbors. Different from them, we dedicate to design an efficient over-sampling algorithm for minority instances, by extending SMOTE to semi-supervised node classification.

\subsection{Graph Neural Network}
In recent years, with the increasing requirements of learning on non-Euclidean space and modeling rich relation information among samples, graph neural networks (GNNs) have received much more attention and are developing rapidly. GNNs generalize convolutional neural networks to graph structured data and have shown great ability in modeling graph structured data. Current GNNs follow a message-passing framework, which is composed of pattern extraction and interaction modeling within each layer~\cite{gilmer2017neural}. Generally, 
existing GNN frameworks can be categorized into two categorizes, i.e., spectral-based~\cite{bruna2013spectral,Tang2019ChebNetEA,Kipf2017SemiSupervisedCW,Hamilton2017InductiveRL} and spatial-based~\cite{duvenaud2015convolutional,atwood2016diffusion}. 

Spectral-based GNNs defines the convolution operation in the Fourier domain by computing the eigendecomposition of the graph Laplacian. Early work~\cite{bruna2013spectral} in this domain involves extensive computation, and is time-consuming. To accelerate, ~\cite{Tang2019ChebNetEA} adopts Chebyshev Polynomials to approximate spectral kernels, and enforces locality constraints by truncating only top-k terms. GCN~\cite{Kipf2017SemiSupervisedCW} takes a further step by preserving only top-2 terms, and obtains a more simplified form. GCN is one of the most widely-used GNN currently. However, all spectral-based GNNs suffer from the generalization problem, as they are dependent on the Laplacian eigenbasis~\cite{Zhou2018GraphNN}. Hence, they are usually applied in the transductive setting, training and testing on the same graph structure. Spatial-based GNNs are more flexible and have stronger in generalization ability. They implement convolutions basing on the neighborhoods of each node. As each node could have different number of neighbors, Duvenaud et al.,~\cite{duvenaud2015convolutional} uses multiple weight matrices, one for each degree. ~\cite{atwood2016diffusion} proposes a diffusion convolution neural network, and ~\cite{niepert2016learning} adopts a fixed number of neighbors for each sample. A more popular model is GraphSage~\cite{Hamilton2017InductiveRL}, which samples and aggregates embedding from local neighbors of each sample. More recently, ~\cite{xu2018powerful} extends expressive power of GNNs to that of WL test, and ~\cite{You2019PositionawareGN} introduce a new GNN layer that can encode node positions.

Despite the success of various GNNs, existing work doesn't consider the class imbalance problem, which widely exists in real-world applications and could significantly reduce the performance of GNNs. Thus, we study a novel problem of synthetic minority oversampling on graphs to facilitate the adoption of GNNs for class imbalance node classification.% In this work, our framework is constructed on top of GraphSage, due to its popularity and efficiency in learning from graphs.

\subsection{Mixup}
Mixup~\cite{zhang2017mixup} is an interpolation-based data augmentation technique designed for computer vision tasks. Through mixing instances of different classes, it is designed to increase the robustness and generalization ability of neural networks. Concretely, it creates new training instances through convex combinations of pairs of examples and their labels, so that neural networks will be regularized to favor simple linear behavior in-between training examples. It works surprisingly well, and is found to increase the robustness against adversarial attacks~\cite{zhang2017mixup}. Later, Manifold Mixup~\cite{verma2019manifold} extends it by performing interpolations in a well-learned embedding space. Recently, Remix~\cite{chou2020remix} extends it to the imbalanced setting by providing a disproportionately higher weight to minority classes during assigning the labels. 

Unlike computer vision domain in which inputs are well-structured and arranged in regular grids, graphs take an irregular structure and nodes are connected. As the mixing of graph topology is not well-defined, and mixed nodes may interfere with each other, it is non-trivial to apply this technique to the graph domain. There have been some attempts addressing these difficulties~\cite{wu2021graphmixup,verma2019graphmix,wang2021mixup}. For example, ~\cite{verma2019graphmix} uses a separate MLP network to conduct mixup and transfer the knowledge to the graph neural network. ~\cite{wang2021mixup} adopts a two-branch graph convolution network to separate the mixup process and clean-training process. In this work, we also works on the graph-structured data, but we do not need to disentangle the mix and learning process. Instead, we use mixup technique to generate new nodes on a learned embedding space, and use an edge predictor to insert them into the existing graph.

%% file: problem_setting.tex
\section{Problem Definition} \label{sec:problem_definition}
In this work, we focus on semi-supervised node classification task on graphs, in the transductive setting. As shown in Figure~\ref{fig:example}, we have a large network of entities, with some labeled for training. Both training and testing are performed on this same graph. Each entity belongs to one class, and the distribution of class sizes are imbalanced. This problem has many practical applications. For example, the under-representation of minority groups on social networks, malicious behavior or fake user accounts which are outnumbered by normal ones, and linked web pages in knowledge base where materials for some topics are limited.

Throughout this paper, we use $\mathcal{G} = \{\mathcal{V}, \mathbf{A}, \mathbf{F}\}$ to denote an attributed network, where $\mathcal{V}=\{v_1,\dots,v_{n}\}$ is a set of $n$ nodes. $\mathbf{A} \in \mathbb{R}^{n \times n}$ is the adjacency matrix of $\mathcal{G}$, and $\mathbf{F} \in \mathbb{R}^{n \times d}$ denotes the node attribute matrix, where $\mathbf{F}[j,:] \in \mathbb{R}^{1 \times d}$ is the node attributes of node j and $d$ is the dimension of the node attributes. $\mathbf{Y} \in \mathbb{R}^{n}$ is the class information for nodes in $\mathcal{G}$. During training, only a subset of $\mathbf{Y}$, $\mathbf{Y}_L$, is available, containing the labels for node subset $\mathcal{V}_L$. There are $m$ classes in total, $\{\mathcal{C}_1, \dots, \mathcal{C}_m\}$. $|\mathcal{C}_i|$ is the size of $i$-th class, referring to the number of samples belong to that class. We use imbalance ratio, $\frac{min_{i}(|\mathcal{C}_i|)}{max_{i}(|\mathcal{C}_i|)}$, to measure the extent of class imbalance. In the imbalanced setting, imbalance ratio of $\mathbf{Y}_L$ is small.

\vspace{0.5em}
\noindent{}\textit{Given $\mathcal{G}$ whose node class set is imbalanced, and labels for a subset of nodes $\mathcal{V}_{L}$, we aim to learn a node classifier $f$ that can work well for both majority and minority classes, i.e.,
\begin{equation}
    f(\mathcal{V}, \mathbf{A}, \mathbf{F}) \rightarrow \mathbf{Y}
\end{equation}
}

%% file: methodology.tex
\section{Augmentation with Minority Nodes}\label{sec:methodology}
\begin{figure}[t!]
  \centering
    \includegraphics[width=0.49\textwidth]{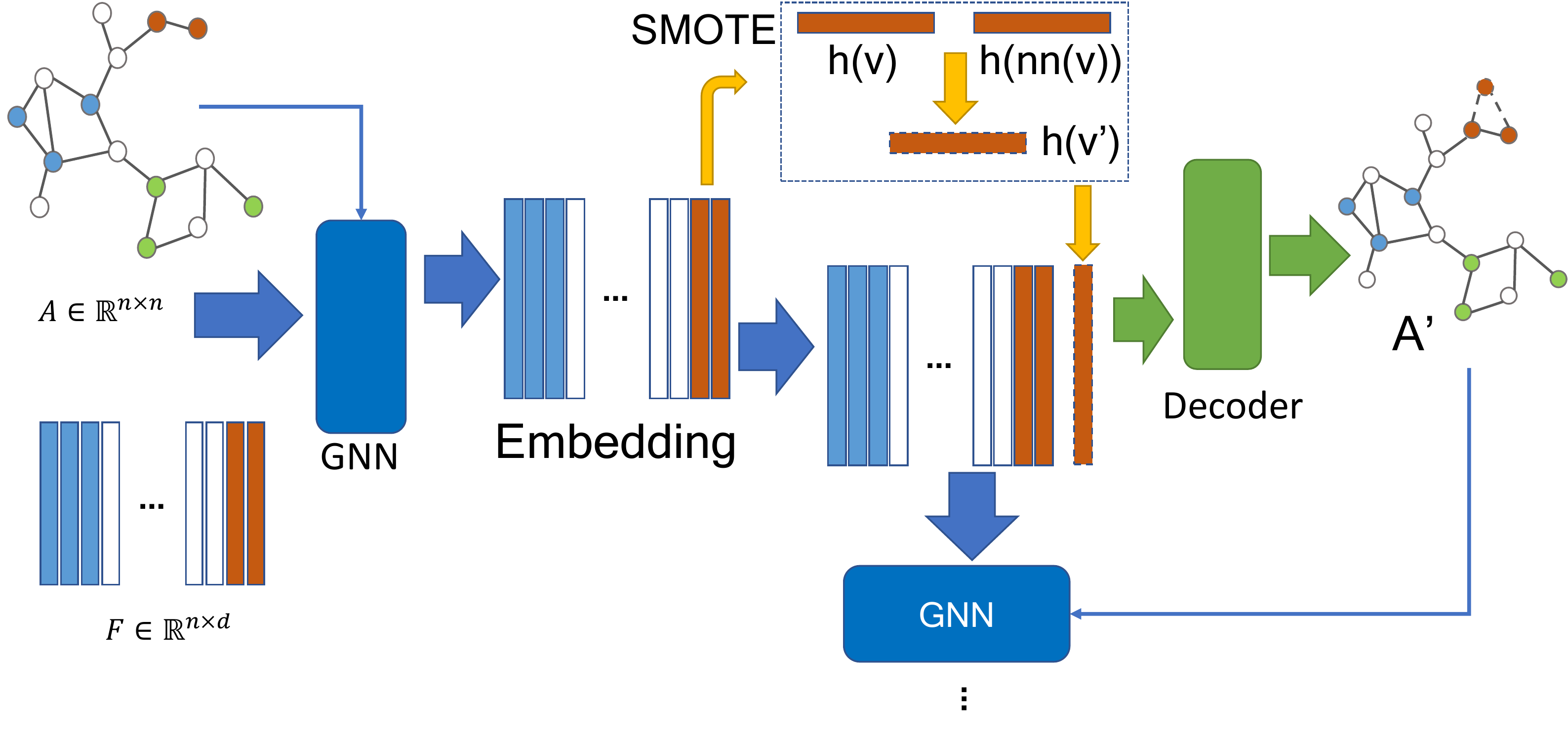}
    \vskip -1em
    \caption{Overview of the {\method} framework, in which pseudo instances of minority classes are generated through interpolation on the learned embedding space. Synthesized new nodes are inserted into the input graph with the help of an edge predictor, and a GNN-based classifier is trained on top of it.} \label{fig:model_architecture}
  \setlength{\abovecaptionskip}{0cm}
\end{figure}

In this section, we give details of synthetic node generation utilizing instances from the minority class. The main idea of {{\method}} is to generate synthetic minority nodes through interpolation in an expressive embedding space acquired by the GNN-based feature extractor, and use an edge generator to predict the links for the synthetic nodes, which forms an augmented balanced graph to facilitate node classification by GNNs. An illustration of the proposed framework is shown in Figure~\ref{fig:model_architecture}. {{\method}} is composed of four components: (i) a GNN-based feature extractor (encoder) which learns node representation that preserves node attributes and graph topology to facilitate the synthetic node generation; (ii) a synthetic node generator which generates synthetic minority nodes in the latent space; (iii) an edge generator which generate links for the synthetic nodes to from an augmented graph with balanced classes; and (iv) a GNN-based classifier which performs node classification based on the augmented graph.  Next, we give the details of each component.

%framework

% Graph neural networks (GNNs) have shown promising results for learning node representations that capture the network topological along with node attributes, and hence are widely used in previous node classification methods. To cope with class imbalance problem of nodes in graph, we propose our over-sampling algorithm, {{\method}}, on top of that structure.

%model design
\subsection{Feature Extractor}
%\suhang{explain why we need feature extractor first: The basic idea of synthetic node generation is to find two similar minority nodes of the same class and do linear interpolation so as to make sure that the generated synthetic nodes also belong to the expected class. This requires us to have good representations of nodes for measuring similarity. Unlike i.i.d data, the similarity of nodes in graphs need to consider both node attributes and graph structures.  Directly generating synthetic nodes cannot fully accurately measure node similarity. Thus, ....} 

One way to generate synthetic minority nodes is to directly apply SMOTE on the raw node feature space. However, this will cause several problems: (i) the raw feature space could be sparse and high-dimensional, which makes it difficult to find two similar nodes of the same class for interpolation; and (ii) it doesn't consider the graph structure, which can result in sub-optimal synthetic nodes. Thus, instead of directly adopting synthetic minority over-sampling in the raw feature space, we introduce a feature extractor learn node representations that can simultaneously capture node properties and graph topology. Generally, the node representations should reflect inter-class and intra-class relations of samples. Similar samples should be closer to each other, and dissimilar samples should be more distant. In this way, when performing interpolation on minority node with its nearest neighbor, the obtained embedding would have a higher probability of representing a new sample belonging to the same minority class. In graphs, the similarity of nodes need to consider node attributes, node labels, as well as local graph structures. Hence, we implement it with GNN, and train it on two down-stream tasks, edge prediction and node classification. %\suhang{train or pretrain?}

The feature extractor can be implemented using any kind of GNNs. In this work, we choose GraphSage as the backbone model structure because it is effective in learning from various types of local topology, and generalizes well to new structures. It has been observed that too deep GNNs often lead to sub-optimal performance, as a result of over-smoothing and over-fitting. Therefore, we adopt only one GraphSage block as the feature extractor.
%We can present the formulations of our adopted GraphSage block here. 
Inside this block, the message passing and fusing process can be written as:
\begin{equation}
\mathbf{h}_{v}^{1}  = \sigma(\mathbf{W}^1 \cdot CONCAT(\mathbf{F}[v,:], \mathbf{A}[v,:] \cdot \mathbf{F})),
\end{equation}
$\mathbf{F}$ represents input node attribute matrix and $\mathbf{F}[v,:]$ represents attribute for node $v$. $\mathbf{A}[v,:]$ is the $v$-th row in adjacency matrix, and $\mathbf{h}_{v}^{1}$ is the obtained embedding for node $v$. $\mathbf{W}^1$ is the weight parameter, and $\sigma$ refers to the activation function such as ReLU. 

\subsection{Synthetic Node Generation}
After obtaining the representation of each node in the embedding space constructed by the feature extractor, now we can perform over-sampling on top of that. We seek to generate realistic representations for new samples from the minority classes. In this work, to perform over-sampling, we adopt the widely used SMOTE algorithm, which augments vanilla over-sampling via changing repetition to interpolation. We choose it due to its popularity, but our framework can also cope with other over-sampling approaches as well. The basic idea of SMOTE is to perform interpolation on samples from the target minority class with their nearest neighbors in the embedding space that belong to the same class. Let $\mathbf{h}_v^1$ be a labeled minority nodes with label as $Y_{v}$. The first step is to find the closest labeled node of the same class as $\mathbf{h}_v^1$, i.e., 
\begin{equation}
    nn(v) = \argmin_{u} \|\mathbf{h}_{u}^{1} - \mathbf{h}_{v}^{1}\|, \quad \text{s.t.}  \quad Y_u = Y_v
\end{equation}
$nn(v)$ refers to the nearest neighbor of $v$ from the same class, measured using Euclidean distance in the embedding space. With the nearest neighbor, we can generate synthetic nodes as
%It can be formalized in the following equation:
\begin{equation}
\begin{aligned}
    \mathbf{h}_{v'}^{1} &= (1-\delta) \cdot \mathbf{h}_{v}^{1} + \delta \cdot \mathbf{h}_{nn(v)}^{1},
\end{aligned}
\end{equation}
where $\delta$ is a random variable, following uniform distribution in the range $[0,1]$. Since $\mathbf{h}_{v}^{1}$  and $\mathbf{h}_{nn(v)}^{1}$ belong to the same class and are very close to each other, the generated synthetic node $\mathbf{h}_{v'}^{1}$ should also belong to the same class. In this way, we can obtain labeled synthetic nodes.
%$nn(v)$ refers to the nearest neighbor of $v$ from the same class, measured using Euclidean distance in the embedding space. $\delta$ is a random variable, following uniform distribution in the range $0 \sim 1$, to perform interpolation. Note that $Y_{v'}^{}$ is set to be the same as $Y_{v}$. 

For each minority class, we can apply SMOTE to generate syntetic nodes. We use a hyper-parameter, over-sampling scale, to control the amount of samples to be generated for each class. Through this generation process, we can make the distribution of class size more balanced, and hence make the trained classifier perform better on those initially under-represented classes.

\subsection{Edge Generator}
%\suhang{why do we want to generate edges. Note that our goal is to facilitate the adoption of GNNs with oversampling. Otherwise, there's no pint of adding edges because we can directly apply classifiers on node embeddings. }

Now we have generated synthetic nodes to balance the class distribution. However, these nodes are isolated from the raw graph $\mathcal{G}$ as they don't have links. % which cannot facilitate the message passing process of a GNN classifier. %Howver, 
Thus, we introduce an edge generator to model the existence of edges among nodes. 
As GNNs need to learn how to extract and propagate features simultaneously, this edge generator can provide relation information for those synthesized samples, and hence facilitate the training of GNN-based classifier. This generator is trained on real nodes and existing edges, and is used to predict neighbor information for those synthetic nodes. These new nodes and edges will be added to the initial adjacency matrix $\mathbf{A}$, and serve as input the the GNN-based classifier.

In order to maintain model's simplicity and make the analysis easier, we adopt a vanilla design, weighted inner production, to implement this edge generator as:
\begin{equation}\label{eq:edge}
\begin{aligned}
    \mathbf{E}_{v,u} &= sigmoid(\sigma((\mathbf{h}_v^1)^T \cdot \mathbf{S} \cdot \mathbf{h}_u^1)),
\end{aligned}
\end{equation}
where $\mathbf{E}_{v,u}$ refers to the predicted relation information between node $v$ and $u$, and $\mathbf{S}$ is the parameter matrix capturing the interaction between nodes. The loss function for training the edge generator is
\begin{equation}
\begin{aligned}
    \mathcal{L}_{edge} = \|\mathbf{E} - \mathbf{A}\|_F^2,
\end{aligned}
\end{equation}
where $\mathbf{E}$ refers to predicted connections between nodes in $\mathcal{V}$, i.e., no synthetic nodes.
Since we learn an edge generator which is good at reconstructing the adjacency matrix using the node representations, it should give good link predictions for synthetic nodes. 

With the edge generator, we attempt two strategies to put the predicted edges for synthetic nodes into the augmented adjacency matrix. %we attempt two strategies.  
In the first strategy, this generator is optimized using only edge reconstruction, and the edges for the synthetic node $v'$ is generated by setting a threshold $\eta$:
\begin{equation}
    \tilde{\mathbf{A}}[v', u] = 
    \begin{cases}
    1, & \text{if } \mathbf{E}_{v', u} > \eta \\
    0,              & \text{otherwise}.
\end{cases}
\end{equation}
where $\tilde{\mathbf{A}}$ is the adjacency matrix after over-sampling, by inserting new nodes and edges into $\mathbf{A}$, and will be sent to the classifier. 

In the second strategy, for synthetic node $v'$, we use soft edges instead of binary ones:
\begin{equation}\label{eq:adj}
    \tilde{\mathbf{A}}[v', u] = \mathbf{E}_{v', u},
\end{equation}
In this case, gradient on $\tilde{\mathbf{A}}$ can be propagated from the classifier, and hence the generator can be optimized using both edge prediction loss and node classification loss, which will be introduced later. Both two strategies are implemented, and their performance are compared in the experiment part.

\subsection{GNN Classifier}
% After over-sampling in the embedding space, the data size of different classes becomes balanced, and an unbiased classifier would be able to be trained on that. 
Let $\tilde{\mathbf{H}}^1$ be the augmented node representation set by concatenating $\mathbf{H}^1$ (embedding of real nodes) with the embedding of the synthetics nodes, and $\tilde{\mathcal{V}}_L$ be the augmented labeled set by incorporating the synthetic nodes into $\mathcal{V}_L$. Now we have an augmented graph 
$ {\tilde{\mathcal{G}}}=\{\tilde{\mathbf{A}},\tilde{\mathbf{H}}\}$ with labeled node set $\tilde{\mathcal{V}}_L$. The data size of different classes in $\tilde{\mathcal{G}}$ becomes balanced, and an unbiased GNN classifier would be able to be trained on that.
% Learning to utilize node embedding and relation information simultaneously, the classifier is also constructed basing on GNN structures. 
Specifically, we adopt another GraphSage block, followed by a linear classification layer for node classification on $\tilde{G}$ as:
\begin{equation}
\mathbf{h}_{v}^{2}  = \sigma(\mathbf{W}^2 \cdot CONCAT(\mathbf{h}_{v}^{1},  \tilde{\mathbf{A}}[v,:] \cdot \tilde{\mathbf{H}}^{1})),
\end{equation}
\begin{equation}\label{eq:predict}
\mathbf{P}_{v}  = softmax(\sigma(\mathbf{W}^c \cdot CONCAT(\mathbf{h}_{v}^{2}, \tilde{\mathbf{A}}[:,v] \cdot \mathbf{H}^{2} ))),
\end{equation}
where $\mathbf{H}^2$ represents node representation matrix of second GraphSage block, and $\mathbf{W}$ refers to the weight parameters. $\mathbf{P}_{v}$ is the probability distribution on class labels for node $v$. The cost fucntion for the classifier module is given as:
\begin{equation}
\mathcal{L}_{node}  = -\sum_{u \in \tilde{\mathcal{V}}_L} \sum_{c}  \mathbf{1}(Y_{u}==c) \cdot log(\mathbf{P}_{v}[c]).
\end{equation}
where $\mathbf{P}_{v}[c]$ is the predicted probability of node $v$ belong to class $c$. $\mathbf{1}(Y_{u}==c)=1$ if $Y_u = c$; otherwise, $\mathbf{1}(Y_{u}==c)=0$. During testing, the predicted class for node $v$, $Y_{v}^{'}$ will be set as the class with highest probability, i.e., $Y_v^{'} = \argmax_{c}\mathbf{P}_{v}[c]$.
% \begin{equation}
% Y_v^{'} = \argmax_{c}\mathbf{P}_{v}[c]
% \end{equation}

%loss design
\subsection{Optimization Objective}
Putting the feature extractor, synthetic node generator, edge generator and GNN classifier together, previous parts together, the final objective function of {{\method}} can be written as:
\begin{equation}
\begin{aligned}
  \min_{\theta, \phi, \varphi} & \mathcal{L}_{node} + \lambda \cdot \mathcal{L}_{edge},
\end{aligned}
\end{equation}
wherein $\theta, \phi, \varphi$ are the parameters for feature extractor, edge generator, and node classifier respectively. As the model's performance is dependent on the quality of embedding space and generated edges, to make training phrase more stable, we also tried pre-training feature extractor and edge generator using $\mathcal{L}_{edge}$. 

The design of {{\method}} has several advantages: (i) it is easy to implement synthetic minority over-sampling process. Through uniting interpolated node embedding and predicted edges, new samples can be generated; (ii) the feature extractor is optimized using training signal from both node classification task and edge prediction task. Therefore, rich intra-class and inter-class relation information would be encoded in the embedding space, making the interpolation more robust; and (iii) it is a general framework. It can cope with different structure choices for each component, and different regularization terms can be enforced to provide prior knowledge.

%\subsection{Discussion}
%make comparisons over different over-sampling approach
%A comparison between the idea of {{\method}} and those of previous over-sampling algorithms is shown in Figure~\ref{fig:comparison}, to illustrate their differences. Vanilla SMOTE generates new samples in the original input domain, which has no natural way to interpolate the edges, and high complexity of input space could easily make interpolation to generate out-of-domain samples. Deep Over-sampling~\cite{ando2017deep} works on the embedding space, which can help to generate in-domain samples, but it can not produce edges for them. As a result, it can not give training signal to the GNN network to guide the message passing among neighbors, which is vital for studying on graphs. {{\method}}, on the other hand, seeks to work on the intermediate embedding space, and generate the presentation for new samples as well as their neighborhood information. Therefore, it can help to guide the training of GNN, and make newly generated samples more realistic at the same time.

% \begin{figure}[t!]
%   \centering
%     \includegraphics[width=0.45\textwidth]{comparison.pdf}
%     \vskip -1em
%     \caption{Comparison of over-sampling approaches.} \label{fig:comparison}
%   \setlength{\abovecaptionskip}{0cm}
% \end{figure}

\section{Augmentation with Majority Nodes}\label{sec:mixup}

In the previous part, we have shown how SMOTE can be extended to the graph domain. The proposed {{\method}} can generate more instances of minority classes, through interpolating in the embedding space and synthesizing relation information. However, this data augmentation strategy only utilizes data from the minority classes. As nodes from the minority classes are usually limited, great potential could lie behind incorporating majority nodes. Therefore in this section, we explore the utilization of majority classes to further improve the graph augmentation process.

Concretely, we explore to generate mixed ``in-between'' nodes through conducting cross-class node interpolations and supervising them with mixed labels, inspired by mixup~\cite{zhang2017mixup} technique. These generated pseudo nodes can help make the classification boundary smoother as well as more discriminative~\cite{wang2021mixup}. This extension faces the same difficulties as interpolation inside the same class. Class distributions are required to be well-concentrated to guarantee the reliability of generated mix labels, and relation information for those generated new nodes needs to be provided. Hence, it can be implemented by extending the {\method} framework. 

To mix nodes of different classes on the graph, we propose a strategy composed of three steps:
\begin{itemize}
    \item \textit{(optional)} Obtain pseudo labels for those nodes without supervision;
    \item Synthesize new instances through interpolating existing nodes on both embedding and label space, and generating edges for them;
    \item Insert synthetic nodes into the graph, and train a GNN model on top of it.
\end{itemize}

Now, we will provide the implementations step by step.

\subsection{Obtain Pseudo Labels}
In the semi-supervised setting, labeled nodes are usually limited in the node classification task, especially for those minority classes. When the labeled set is small, generated new nodes may not reflect real distributions well, and pseudo labels from mixup may become unreliable. Hence, we adopt an optional pseudo label obtaining process, to create artificial supervisions for those unlabeled nodes. 

Concretely, in each update step, we first use predictions of the intermediate trained classifier as the pseudo labels $\hat{Y}$:
\begin{equation}
    \hat{Y}_{v} = \begin{cases}
    Y_{v}, \quad v \in \mathcal{V}_L \\
    \argmax_{c}\mathbf{P}_{v}[c], \quad v \notin \mathcal{V}_L \\
    \end{cases} 
\end{equation}
where $\mathcal{V}_L$ is the labeled node set. $\mathbf{P}_{v}[c]$ is the predicted probability of node $v$ being class $c$. As the predicted labels are inaccurate, we set a threshold $\mathcal{T}$ to filter out nodes with little confidence, ie., $\max_{c}\mathbf{P}_{v}[c] \leq \mathcal{T}$. 

The obtained node set can be used to go through the following data augmentation procedures and synthesize new mixed nodes. %As this process may introduce noises and its effect is dependent upon the trained model as well as the dataset, we take this step as optional. 
Performance comparison of this extension will be tested on real-world datasets in the experiment part. 

\subsection{Mixup on Graph}
In this subsection, we show how new nodes can be generated through interpolation across classes. The model framework is the same as introduced in the previous section, with the feature extractor, edge generator and GNN-based classifier introduced in Section~\ref{sec:methodology}. The only different part is the node synthesis process.

To perform mixup on graph, we generate new nodes through interpolation in both the embedding space and node label space. For a labeled node $v$ from minority classes, we randomly select a node $u$ form the majority classes and generate synthetic labeled node $\hat{v}$ as:
\begin{equation} \label{eq:mix}
\begin{aligned}
\begin{cases}
    \mathbf{h}_{\hat{v}}^{1} &= (1-\delta') \cdot \mathbf{h}_{v}^{1} + \delta' \cdot \mathbf{h}_{u}^{1}, \\
    \hat{Y}_{\hat{v}} &= (1-\delta') \cdot \hat{Y}_{v} + \delta' \cdot \hat{Y}_{u},
\end{cases} \\
\end{aligned}
\end{equation}
\begin{equation}
\text{s.t.} \quad  \hat{Y}_v \neq \hat{Y}_u
\end{equation}
where $\mathbf{h}_{v}^{1}$ is the embedding of node $v$ from the feature extractor, and $\delta' \sim \bigcup(0, b)$ is a randomly generated variable following uniform distribution within the scale $[0,b]$ for conducting interpolation.  This interpolation process can synthesize ``in-between'' samples, and their labels are also set as ``in-between''. %However, these heuristically obtained labels could be unreliable, as selected $v$ or $u$ may not lie on the distribution boundaries, resulting in generated label being biased~\cite{han2005borderline}. In this task, classes are imbalanced with only few samples of minority classes. Hence, this bias could be misleading and particularly harmful towards those minority classes. With these observations, we select a small $b$ as $0.5$. 
As observed in ~\cite{han2005borderline,chou2020remix}, it is difficult to guarantee the alignment between mixed features and mixed labels when classes are imbalanced. To this end, we adopt a small interpolation scale $b$ of $0.5$, since it would generate nodes closer to the minority ones, and reduce the biases in heuristically set labels.  An analysis on $b$ is conducted in Section~\ref{sec:expmixup}. These generated mixed nodes provide auxiliary signals for learning class boundaries by filling in the distribution gap among classes and encouraging linear behavior in-between training examples. What is more, more diverse synthetic nodes can be obtained with this extension, as training examples from majority classes are also incorporated in the generation process.

The next step is to provide edges for generated nodes. Edges are discrete, and it is difficult to heuristically set them for these ``in-between'' samples as ``in-between'' edges make little sense. Addressing this, we train an edge predictor to model the existence probability between node pairs, and apply it to produce relation information. For mixed node set $\hat{\mathcal{V}}$, this process follows Equation \ref{eq:edge}-\ref{eq:adj}. After inserting them to the input graph, we annotate the augmented graph as ${\tilde{\mathcal{G}}}$. With these generated ``in-between'' nodes being included into ${\tilde{\mathcal{G}}}$, the distribution boundary among classes will be stressed, which provides an auxiliary signal for the classifier.

\subsection{Optimization Objective}
After conducting mixup and inserting mixed new nodes into the existing graph, now we can update the node classifier on this augmented graph ${\tilde{\mathcal{G}}}$, with auxiliary training signals from these synthetic nodes. We use $\mathbf{P}_{\hat{v}}$ to represent the predicted class distribution of mixed node $\hat{v}$, which is calculated by the GNN-based classifier following Equation~\ref{eq:predict}. It can be supervised using mixed labels $\hat{\mathbf{Y}}$. The optimization objective on mixed nodes is given as:
\begin{equation}
\begin{aligned}
\mathcal{L}_{mix}  &= \mathbb{E}_{v \in \hat{\mathcal{V}}} \mathbb{E}_{u \in \hat{\mathcal{V}}} \mathbb{E}_{\delta' \sim \bigcup(0,b)} l \big(\mathbf{P}_{\hat{v}}, \hat{Y}_{\hat{v}}  \big), \\
\text{s.t.} & \quad  \hat{Y}_v \neq \hat{Y}_u.
\end{aligned}
\end{equation}
where $l$ represents the loss function like cross entropy, and $\hat{v}$ is generated following Equation ~\ref{eq:mix}.

\begin{equation}
\begin{aligned}
  \min_{\theta, \phi, \varphi} & \mathcal{L}_{node} + \lambda \cdot \mathcal{L}_{edge} + \lambda_2 \cdot \mathcal{L}_{mix},
\end{aligned}
\end{equation}

\subsection{Training Algorithm}
The full pipeline of running our framework can be summarized in Algorithm 1. Inside each optimization step, we first obtain node representations using the feature extractor in line $6$. Then, from line $7$ to line $11$, we perform over-sampling in the embedding space to make node classes balanced. If mixup is required, from line $13$ to line $14$ we conduct interpolation across node classes to obtain mixed nodes. After predicting edges for generated new samples in line $16$, the following node classifier can be trained on top of that over-sampled graph. The full framework is trained altogether with edge prediction loss and node classification loss, as shown in line $18$ and line $20$.

\begin{algorithm}
  \caption{Full Training Algorithm}
  \label{alg:Framwork}
  %\scalebox{0.75}{
  \begin{algorithmic}[1] 
  \REQUIRE %??????????Input
    $\mathcal{G} = \{\mathcal{V}, \mathbf{A}, \mathbf{F}, \mathbf{Y}\} $, over-sampling scale, $\lambda$, $\lambda_2$
  \ENSURE $\text{Predicted node class } \mathbf{Y}^{'}$
    \STATE Randomly initialize the feature extractor, edge generator and node classifier;
    \IF{Require pre-train}
    \STATE Fix other parts, train the feature extractor and edge generator module until convergence, based on loss $\mathcal{L}_{edge}$;
    \ENDIF
    \WHILE {Not Converged}
    \STATE Input $\mathcal{G}$ to feature extractor, obtaining $\mathbf{H}^{1}$; 
    \FOR{class c in minority classes}
    \FOR{$i$ in $size(c) \cdot \text{over-sampling scale}$}
    \STATE Generate a new sample in class c, Following Equation (3) and (4);
    \ENDFOR
    \ENDFOR
    \IF {mixup}
        \STATE ({\textit{Optional}}) Obtain pseudo labels for unsupervised nodes following Equation (14);
        \STATE Generate mixed nodes following Equation (15);
    \ENDIF
    \STATE Generate $\mathbf{A}^{'}$ using edge generator, basing on Equation (7) or (8);
    \IF {mixup}
        \STATE Update the model using $\mathcal{L}_{node} + \lambda \cdot \mathcal{L}_{edge} + \lambda_2 \cdot \mathcal{L}_{mix}$; 
    \ELSE
        \STATE Update the model using $\mathcal{L}_{node} + \lambda \cdot \mathcal{L}_{edge}$;
    \ENDIF
    \ENDWHILE
    \RETURN Trained feature extractor, edge predictor, and node classifier module.
  \end{algorithmic}
  %}%% resizebox
\end{algorithm}

%% file: experiment.tex
\section{Experiments} \label{sec:experiments}
In this section, we conduct experiments to evaluate the benefits of proposed method for the node classification task when classes are imbalanced. Both artificial and genuine imbalanced datasets are used, and different configurations are adopted to test its generalization ability. Particularly, we want to answer the following questions:
\begin{itemize}
    \item How effective is {\method} in imbalanced node classification task?
    \item How different choices of over-sampling scales would affect the performance of {\method}?
    \item Can {\method} generalize well to different imbalance ratios, or different base model structures?
    \item How would mixup help in the case of extreme imbalance ratios, and in the few-label scenario?
\end{itemize}
We begin by introducing the experimental settings, including datasets, baselines, and evaluation metrics. We then conduct experiments to answer these questions.

\subsection{Experimental Settings}
\subsubsection{Datasets}
We conduct experiments on two widely used publicly available datasets for node classification, Cora~\cite{Sen2008CollectiveCI} and BlogCatalog~\cite{tang2009relational}, and one fake account detection dataset, Twitter~\cite{mohammadrezaei2018identifying}. The details of these three datasets are given as follows:
\begin{itemize}[leftmargin=*]
    \item \textbf{Cora}: Cora is a citation network dataset for transductive learning setting. It contains one single large graph with $2,708$ papers from $7$ areas. Each node has a $1433$-dim attribution vector, and a total number of $5,429$ citation links exist in that graph. In this dataset, class distributions are relatively balanced, so we use an imitative imbalanced setting: three random classes are selected as minority classes and down-sampled. All majority classes have a training set of $20$ nodes. For each minority class, the number is $20 \times imbalance\_ratio$. We vary $imbalance\_ratio$ to analyze the performance of {\method} under various imbalanced scenarios.
    \item \textbf{BlogCatalog}: This is a social network dataset crawled from BlogCatalog\footnote{http://www.blogcatalog.com}, with $10,312$ bloggers from $38$ classes and $333,983$ friendship edges. The dataset doesn't contain node attributes. Following ~\cite{Perozzi2014DeepWalkOL}, we attribute each node with a $64$-dim embedding vector obtained from Deepwalk. Classes in this dataset follow a genuine imbalanced distribution, with $14$ classes smaller than $100$, and $8$ classes larger than $500$. For this dataset, we use $25\%$ samples of each class for training and 25\% for validation, the remaining $50\%$ for testing.
    \item \textbf{Twitter}: This dataset is crawled by ~\cite{mohammadrezaei2018identifying} with a dedicated API crawler from Twitter\footnote{https://twitter.com} on bot infestation problem. It has $5,384,160$ users in total. Among them, $63,167$ users are bots. In this work, we split a connected sub-graph from it containing $61,122$ genuine users and $2,045$ robots. Node embedding is obtained through Deepwalk, appended with node degrees. This dataset is used for binary classification. The imbalance ratio is roughly $1:30$. We randomly select $25\%$ of total samples for training, 25\% for validation, and the remaining $50\%$ for testing.
\end{itemize}
%\suhang{add another real-world imbalanced dataset such as malicious account detection dataset}

\subsubsection{Baselines}
We compare {\method} with representative and state-of-the-art approaches for handling imbalanced class distribution problem, which includes:
\begin{itemize}[leftmargin=*]
    \item Over-sampling: A classical approach for imbalanced learning problem, by repeating samples from minority classes. We implement it in the raw input space, by duplicating $n_s$ minority nodes along their edges. In each training iteration, $\mathcal{V}$ is over-sampled to contain $n+n_s$ nodes, and $\mathbf{A} \in \mathbb{R}^{(n+n_s) \times (n+n_s)}$.
    \item Re-weight~\cite{Yuan2012SamplingR}: This is a cost-sensitive approach which gives class-specific loss weight. It assigns higher loss weights to samples from minority so as to alleviate the issue of majority classes dominating the loss function.
    \item SMOTE~\cite{chawla2002smote}: Synthetic minority oversampling techniques generate synthetic minority samples by interpolating a minority samples and its nearest neighbors of the same class. For newly generated nodes, its edges are set to be the same as the target node.  
    \item Embed-SMOTE~\cite{ando2017deep}: An extension of SMOTE for deep learning scenario, which perform over-sampling in the intermediate embedding layer instead of the input domain. We set it as the output of last GNN layer, so that there is no need to generate edges.
    \item RECT~\cite{wang2020network}: It proposes two regularization terms on top of learned class-level semantic embeddings, to address imbalanced learning in the extreme scenario.
    \item ReNode~\cite{chen2021topology}: A topology-aware re-weighting method, which re-weights labeled nodes based on influence received by their neighbors.
    \item DRGCN~\cite{shi2020multi}: It tackles the class imbalance problem by encouraging the separation between classes in the latent embedding space with the adversarial training paradigm.
\end{itemize}

Basing on the strategy for training edge generator and setting edges, four variants of {\method} are tested:
\begin{itemize}[leftmargin=*]
    \item ${\method}_T$: The edge generator is trained using loss from only edge prediction task. The predicted edges are set to binary values with a threshold before sending to GNN-based classifier;
    \item ${\method}_O$: Predicted edges are set as continuous so that gradient can be calculated and propagated from GNN-based classifier. The edge generator is trained along with other components with training signals from both edge generation task and node classification task;
    \item ${\method}_{preT}$: An extension of ${\method}_T$, in which the feature extractor and edge generator are pre-trained on the edge prediction task, before fine-tuning on Equation.13. During fine-tuning, edge generator is optimized using only $\mathcal{L}_{edges}$;
    \item ${\method}_{preO}$: An extension of ${\method}_O$, in which a pre-training process is also conducted before fine-tuning, same as ${\method}_{preT}$.
\end{itemize}
%\suhang{Tianxiang, need to rename these variants, say use $${\method}_T$$. {\method} Thresh is very confusing when put in the content as Thresh is isolated from {\method}. In addition, it is too long.}

In the experiments, all these methods are implemented and tested on the same GNN-based network for a fair comparison.

\begin{table*}[t]\scriptsize
  \setlength{\tabcolsep}{4.5pt}
  
  \caption{Comparison of different approaches for imbalanced node classification.}\label{tab:result} 
  \vskip -1em
  \begin{tabular}{p{2.4cm} | p{1.12cm}  p{1.52cm}  p{1.52cm} | p{1.12cm}  p{1.52cm}  p{1.52cm} | p{1.12cm}  p{1.52cm}  p{1.52cm} }

    \hline
     &  \multicolumn{3}{|c|}{Cora} &  \multicolumn{3}{|c|}{BlogCatalog} &  \multicolumn{3}{|c}{Twitter} \\
    \hline
    Methods & ACC & AUC-ROC & F Score & ACC & AUC-ROC & F Score & ACC & AUC-ROC & F Score \\
    \hline
    Origin & $68.1\pm0.1$ & $0.914\pm0.002$ & $0.684\pm0.003$ & $21.0\pm0.4$ & $0.586\pm0.002$ & $0.074\pm0.002$ & $96.7\pm0.4$ & $0.577\pm0.003$ & $0.494\pm0.001$  \\
    over-sampling & $69.2\pm0.9$ & $0.918\pm0.005$ & $0.666\pm0.008$ & $20.3\pm0.4$ & $0.599\pm0.003$ & $0.077\pm0.001$ & $91.3\pm0.6$ & $0.601\pm0.011$ & $0.513\pm0.003$  \\
    Re-weight & $69.7\pm0.8$ & $0.928\pm0.005$ & $0.684\pm0.004$ & $20.6\pm0.5$ & $0.587\pm0.003$ & $0.075\pm0.003$ & $91.5\pm0.5$ & $0.603\pm0.004$ & $0.515\pm0.002$  \\
    SMOTE & $69.6\pm1.1$ & $0.920\pm0.008$ & $0.673\pm0.003$ & $20.5\pm0.4$ & $0.595\pm0.003$ & $0.077\pm0.001$ & $91.4\pm0.5$ & $0.604\pm0.007$ & $0.514\pm0.002$  \\
    Embed-SMOTE & $68.3\pm0.7$ & $0.913\pm0.002$ & $0.673\pm0.002$ & $20.5\pm0.3$ & $0.588\pm0.002$ & $0.076\pm0.001$ & $94.3\pm0.4$ & $0.606\pm0.005$ & $0.514\pm0.002$  \\
    ReNode & $68.5\pm0.2$ & $0.916\pm0.001$ & $0.689\pm0.003$ & $21.1\pm0.6$ & $0.592\pm0.003$ & $0.076\pm0.002$ & $\underline{96.5}\pm0.5$ & $0.583\pm0.004$ & $0.503\pm0.002$ \\
    RECT & $68.5\pm1.3$ & $0.921\pm0.007$ & $0.689\pm0.006$ & $20.2\pm0.7$ & $0.593\pm0.004$ & $0.073\pm0.003$ & $90.9\pm0.7$ & $0.605\pm0.013$ & $0.509\pm0.006$ \\
    DRGCN & $69.4\pm1.1$ & $0.932\pm0.006$ & $0.691\pm0.007$ & $20.8\pm0.6$ & $0.603\pm0.005$ & $0.078\pm0.004$ & $92.7\pm0.6$ & $0.608\pm0.011$ & $0.516\pm0.008$ \\
    \hline
    ${\method}_T$ & $71.3\pm0.8$ & $0.929\pm0.006$ & $0.720\pm0.002$ & $20.6\pm0.5$ & $0.602\pm0.004$ & $0.083\pm0.003$ & $92.9\pm0.5$ & $0.622\pm0.003$ & $0.519\pm0.001$ \\
    ${\method}_O$ & $70.9\pm1.0$ & $0.927\pm0.011$ & $0.712\pm0.003$ & $21.5\pm1.0$ & $0.591\pm0.012$ & $0.080\pm0.005$ & $90.5\pm0.8$ & $0.616\pm0.006$ & $0.515\pm0.003$ \\
    ${\method}_{preT}$ & $72.7\pm0.3$ & $0.931\pm0.002$ & $0.726\pm0.001$  & $\underline{24.9}\pm0.2$ & $\underline{0.641}\pm0.001$ & $\underline{0.126}\pm0.001$ & $93.7\pm0.3$ & $\underline{0.639}\pm0.002$ & $0.531\pm0.001$ \\
    ${\method}_{preO}$ & $\underline{73.6}\pm0.1$ & $\underline{0.934}\pm0.002$ & $\underline{0.727}\pm0.001$ & $24.3\pm0.2$ & $\underline{0.641}\pm0.002$ & $0.123\pm0.001$ & $94.1\pm0.2$ & $0.636\pm0.001$ & $\underline{0.532}\pm0.001$ \\
    \hline
  \end{tabular}
\end{table*}

\subsubsection{Evaluation Metrics}
%\suhang{following existing work to evaluate imbalanced classification, we adopt... also cite papers}\
Following existing works in evaluating imbalanced classification~\cite{rout2018handling,johnson2019survey}, we adopt three criteria: classification accuracy(ACC), mean AUC-ROC score~\cite{bradley1997use}, and mean F-measure. ACC is computed on all testing examples at once, therefore may underweight those under-represented classes. AUC-ROC score illustrates the probability that the corrected class is ranked higher than other classes, and F-measure gives the harmonic mean of precision and recall for each class. Both AUC-ROC score and F-measure are calculated separately for each class and then non-weighted average over them, therefore can better reflect the performance on minority classes. %\suhang{how do you average? is it weighted average or non-weighted average, generally, we don't mention average. If you mention, you need to make it clear.}

\subsubsection{Configurations}
All experiments are conducted on a $64$-bit machine with Nvidia GPU (Tesla V100, 1246MHz , 16 GB memory), and ADAM optimization algorithm is used to train the models.

For all methods, the learning rate is initialized to $0.001$, with weight decay being $5e-4$. $\lambda$ is set as $1e-6$, since we did not normalize $\mathcal{L}_{edge}$ and it is much larger than $\mathcal{L}_{node}$. On Cora dataset, imbalance\_ratio is set to $0.5$ and over-sampling scale is set as $2.0$ if not specified otherwise. For BlogCatalog and Twitter dataset, imbalance\_ratio is not involved, and over-sampling scale is set class-wise: $\frac{n}{m \cdot |\mathcal{C}_i|}$ for minority class $i$, to make the class size balanced. Besides, all models are trained until converging, with the maximum training epoch being $5000$.

\subsection{Imbalanced Classification Performance}
To answer the first question, we compare the imbalanced node classification performance of {\method} with the baselines on aforementioned three datasets. Each experiment is conducted 3 times to alleviate the randomness. The average results with standard deviation are reported in Table~\ref{tab:result}. From the table, we can make following observations:
\begin{itemize}
    \item All four variants of {\method} showed significant improvements on imbalanced node classification task, compared to the ``Origin'' setting, in which no special algorithm is adopted. They also outperform almost all baselines in all datasets, on all evaluation metrics. These results validate the effectiveness of proposed framework.
    \item The improvements brought by {\method} are much larger than directly applying previous over-sampling algorithms. For example, compared with Over-sampling ${\method}_T$ shows an improvement of $0.011, 0.003, 0.021$ in AUC-ROC score, and an improvement of $0.016, 0.014,  0.016$ in AUC-ROC score compared with Embed-SMOTE. This result validates the advantages of {\method} over previous algorithms, in constructing an embedding space for interpolation and provide relation information.
    \item Among different variants of {\method}, pre-trained implementations show much stronger performance than not pre-trained ones. This result implies the importance of a better embedding space in which the similarities among samples are well encoded.
\end{itemize}

To summarize, these results prove the advantages of introducing over-sampling algorithm for imbalanced node classification task. They also validate that {\method} can generate more realistic samples and the importance of providing relation information.

\begin{figure}[t!]
  \centering
    \includegraphics[width=0.45\textwidth]{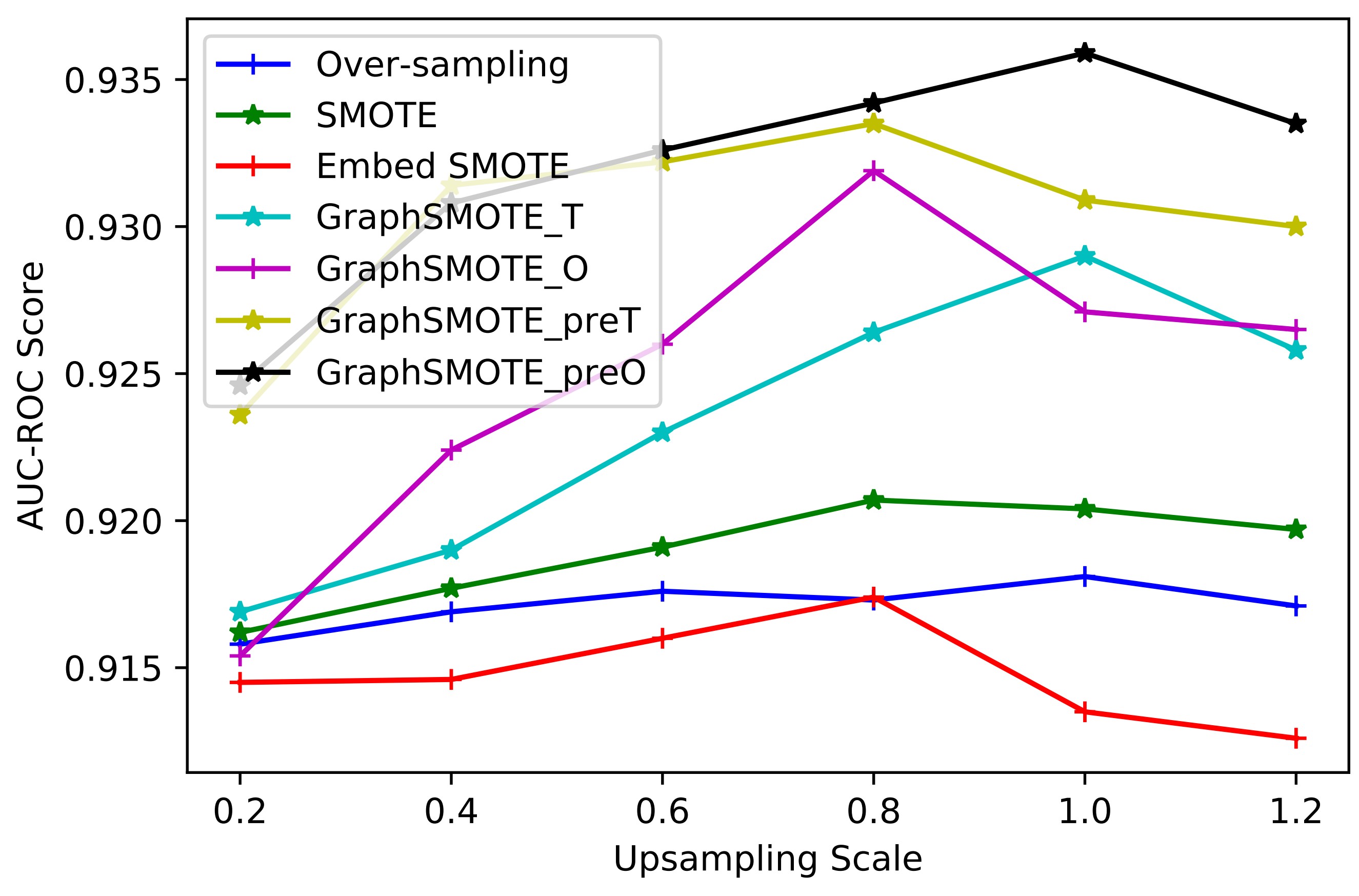}
    \vskip -1em
    \caption{Auc-ROC score achieved by different over-sampling approaches with different upsampling scales. Lower scale means fewer nodes are generated, and higher scale means that more pseudo nodes are synthesized.} \label{fig:UpScale}
  \setlength{\abovecaptionskip}{0cm}
  % \vskip -1em
\end{figure}

\subsection{Influence of Over-sampling Scale}
In this subsection, we analyze the performance change of different algorithms w.r.t different over-sampling scales, in the pursuit of answering the second question. To conduct experiments in a constrained setting, we use Cora dataset and fix imbalance ratio as $0.5$. Over-sampling scale is varied as $\{0.2, 0.4, 0.6, 0.8, 1.0, 1.2\}$. Every experiment is conducted 3 times and the average results are presented in Figure~\ref{fig:UpScale}. From the figure, we make the following observations:
\begin{itemize}
    \item When over-sampling scale is smaller than $0.8$, generating more samples for minority classes, i.e., making the classes more balanced, would help the classifier to achieve better performance, which is as expected because these synthetic nodes not only balance the datasets but also introduce new supervision for training a better GNN classifier.
    \item When the over-sampling scale becomes larger, keeping increasing it may result in opposite effects. It can be observed that the performance remains similar, or degrade a little when changing over-sampling scale from $1.0$ to $1.2$. This is because when too many synthetic nodes are generated, some of these synthetic nodes contain similar/redundant information which cannot further help learn a better GNN.
    \item Based on these observations, generally setting the over-sampling scale set a value that can make the class balanced is a good choice, which is consistent with existing work for synthetic minority oversampling~\cite{buda2018systematic}.
\end{itemize}

\subsection{Influence of Imbalance Ratio}
In this subsection, we analyze the performance of different algorithms with respect to different imbalance ratios, to evaluate their robustness. Experiment is also conducted in a well-constrained setting on Cora, by fixing over-sampling scale to $1.0$, and varying imbalance ratio as $\{0.1,0.2,0.4,0.6\}$. Each experiments are conducted 3 times and the average results are shown in Table~\ref{tab:ImRatio}. From the table, we make the following observations:
\begin{itemize}
    \item The proposed framework {\method} generalizes well to different imbalance ratios. It achieves the best performance across all the settings, which shows the effectiveness of the proposed framework under various scenarios.
    \item The improvement of {\method} is more significant when the imbalance extent is more extreme. For example, when imbalance ratio is $0.1$, ${\method}_{preO}$ outperforms Re-weight by $0.0326$, and the gap reduces to $0.0060$ when the imbalance ratio become $0.6$. This is because when the datasets is not that imbalanced, minority oversampling is not that important, which makes the improvement of proposed algorithm over others not that significant.
    \item Pre-training is important when the imbalance ratio is extreme. When imbalance ratio is $0.1$, ${\method}_{preO}$ shows an improvement of $0.0268$ over ${\method}_{preO}$, and the gap reduces to $0.0055$ when the imbalance ratio changes to $0.6$.
\end{itemize}

\begin{table}[t!]
  \setlength{\tabcolsep}{4.5pt}
  \small
  \caption{Node classification performance in terms of AUC on Cora under various imbalance ratios.} \label{tab:ImRatio}
  \vskip -1em
  \begin{tabular}{c || c | c | c | c  }
    \hline
     &  \multicolumn{4}{|c}{Imbalance Ratio} \\
    \hline
    Methods & $0.1$ & $0.2$ & $0.4$ & $0.6$ \\
    \hline
    Origin & $0.8681$ & $0.8998$ & $0.9139$ & $0.9146$ \\
    over-sampling & $0.8707$ & $0.9039$ & $0.9137$ & $0.9215$ \\
    Re-weight & $0.8791$ & $0.8881$ & $0.9257$ & $0.9306$ \\
    SMOTE & $0.8742$ & $0.9027$ & $0.9161$ & $0.9237$ \\
    Embed-SMOTE & $0.8651$ & $0.8967$ & $0.9188$ & $0.9212$ \\
    \hline
    ${\method}_T$ & $0.8824$ & $\mathbf{0.9162}$ & $0.9262$ & $0.9309$ \\
    ${\method}_O$ & $0.8849$ & $0.9061$ & $0.9216$ & $0.9311$ \\
    ${\method}_{preT}$ & $\mathbf{0.9167}$ & $0.9130$ & $0.9303$ & $0.9317$ \\
    ${\method}_{preO}$ & $0.9117$ & $0.9116$  & $\mathbf{0.9389}$ & $\mathbf{0.9366}$ \\
    \hline
  \end{tabular}
\end{table}

\subsection{Influence of Base Model}
In this subsection, we test generalization ability of the proposed algorithm by applying it to another widely-used graph neural network: GCN. Comparison between it and baselines is presented in Table~\ref{tab:GCN}. All methods are implemented on the same network. Experiments are performed on Cora, with imbalance ratio set as $0.5$ and over-sampling scale as $2.0$. Experiments are run three times, with both averaged results and standard deviation reported. From the result, it can be observed that: 
\begin{itemize}
    \item Generally, {\method} adapt well to GCN-based model. Four variants of it all work well and achieve the best performance, as shown in Table~\ref{tab:GCN}.
    \item Compared with using GraphSage as base model, a main difference is that pre-training seems to be less necessary in this case. We think it may be caused by the fact that GCN is less powerful than GraphSage in representation ability. GraphSage is more flexible and can model more complex relation information, and hence is more difficult to train. Therefore, it can benefit more from obtaining a well-trained embedding space in advance.
\end{itemize}

\begin{table}[h!]
  \scriptsize
  \setlength{\tabcolsep}{4.5pt}
  \caption{Evaluation of different algorithm's performance when changed to GCN as base model.} \label{tab:GCN}
  \vskip -1em
  
  \begin{tabular}{c || c | c | c   }
    \hline
     &  \multicolumn{3}{|c}{Cora} \\
    \hline
    Methods & ACC & AUC-ROC & F Score  \\
    \hline
    Origin & $68.5\pm0.2$ & $0.907\pm0.003$ & $0.663\pm0.001$ \\
    over-sampling & $68.2\pm0.5$ & $0.907\pm0.003$ & $0.665\pm0.003$ \\
    Re-weight & $68.4\pm0.5$ & $0.913\pm0.004$ & $0.672\pm0.002$ \\
    SMOTE & $68.4\pm0.6$ & $0.910\pm0.005$ & $0.665\pm0.003$ \\
    Embed-SMOTE & $69.1\pm0.2$ & $0.910\pm0.003$ & $0.667\pm0.002$  \\
    \hline
    ${\method}_T$ & $69.5\pm0.5$ & $\mathbf{0.920\pm0.003}$ & $0.690\pm0.002$  \\
    ${\method}_O$ & $69.3\pm0.5$ & $0.920\pm0.005$ & $\mathbf{0.707}\pm0.003$  \\
    ${\method}_{preT}$ & $68.8\pm0.1$ & $0.919\pm0.002$ & $0.682\pm0.001$ \\
    ${\method}_{preO}$ & $\mathbf{69.9}\pm0.2$ & $0.914\pm0.002$  & $0.702\pm0.001$  \\
    \hline
  \end{tabular}
\end{table}

%\suhang{do we have hyperparameters that we need to conduct parameter sensitivity analysis}
\subsection{Parameter Sensitivity Analysis}
\begin{figure}[t!]
  \centering
  \subfigure[AUC-ROC Score]{
		\label{fig:auc-lambda}
		\includegraphics[width=0.23\textwidth]{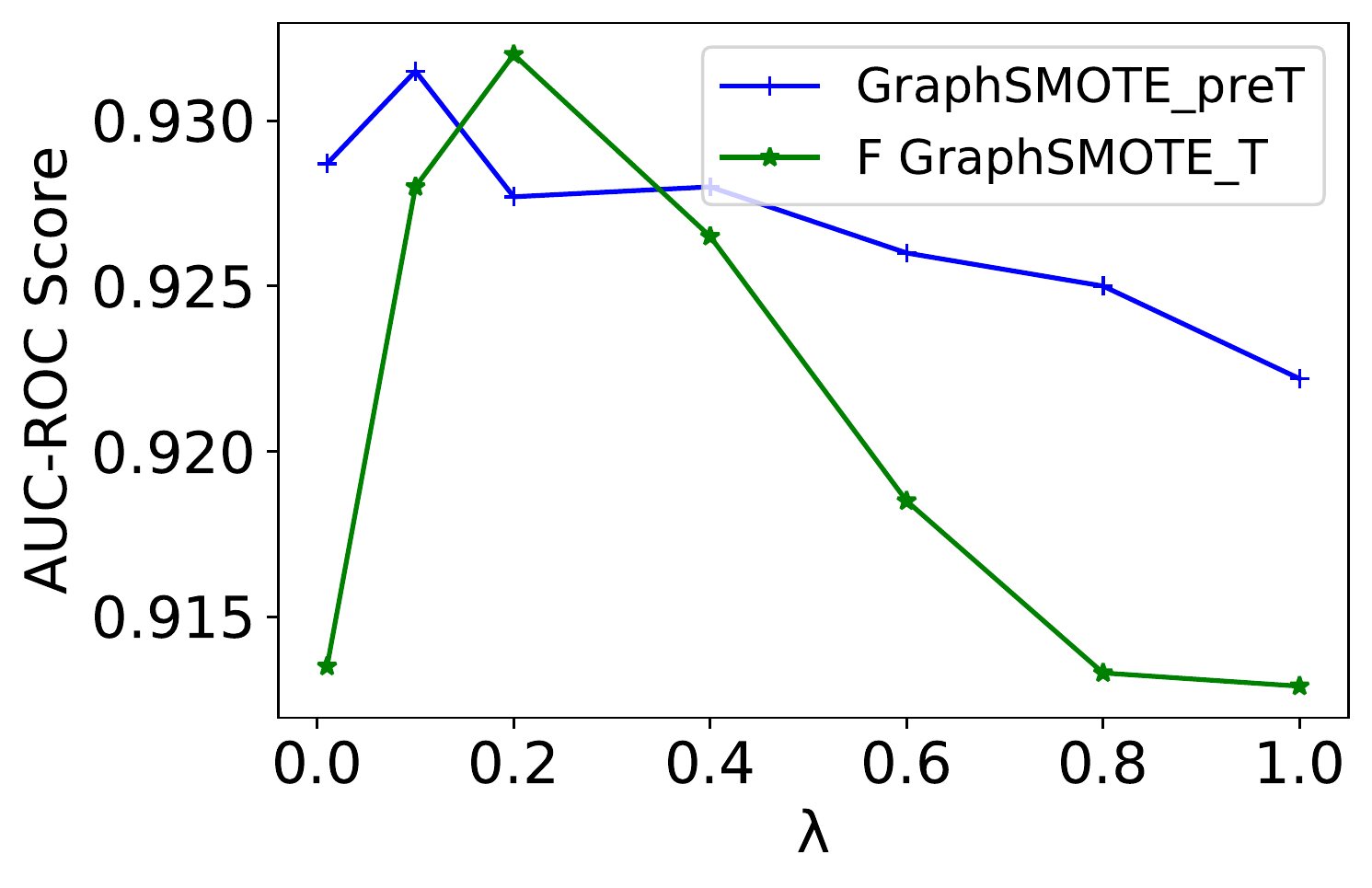}}
    \subfigure[F Measurement]{
		\label{fig:f-lambda}
		\includegraphics[width=0.23\textwidth]{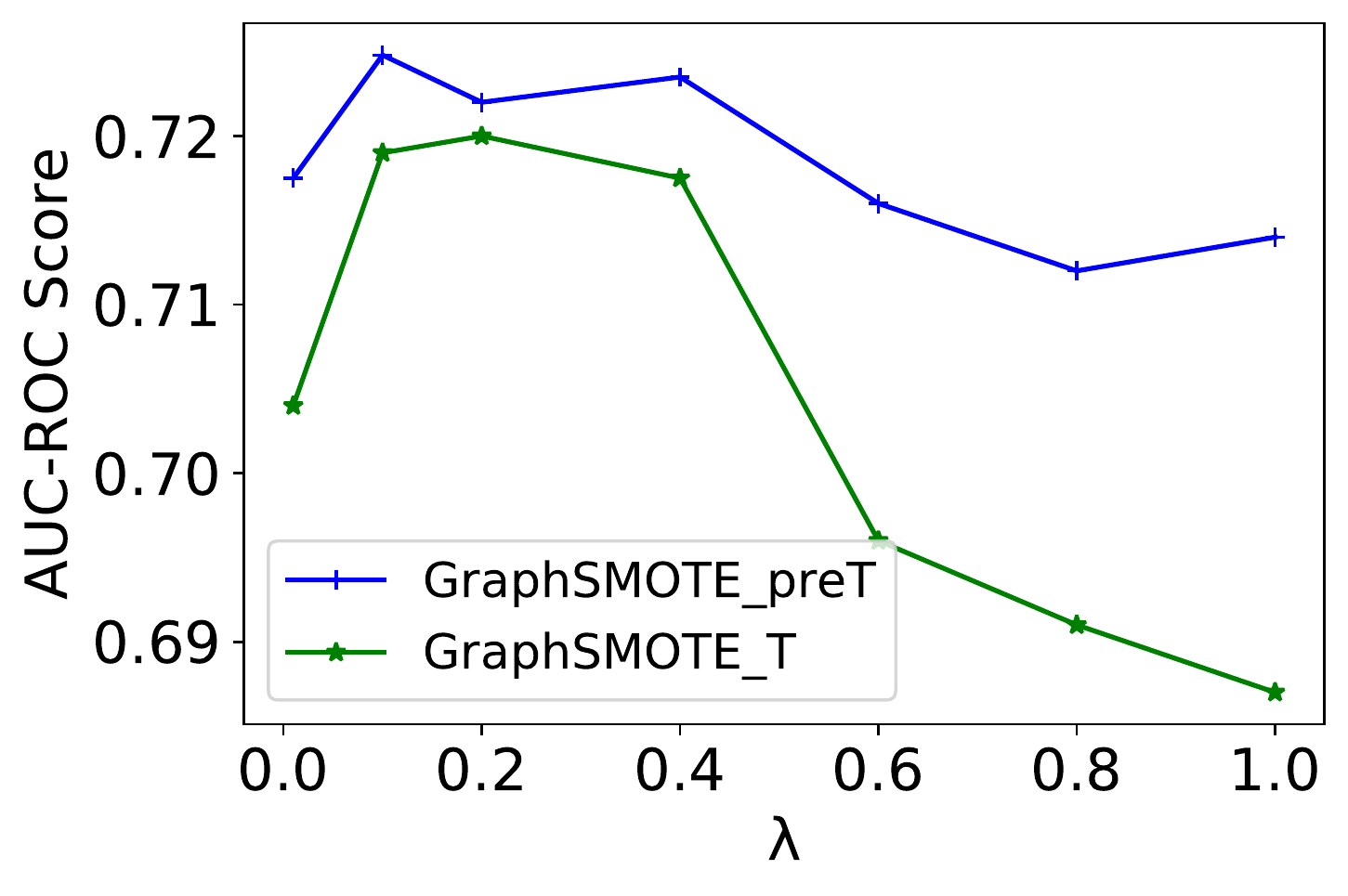}}
		
    \vskip -1em
    \caption{Sensitivity analysis of {\method} with respect to $\lambda$, the weight of edge prediction loss. Performances on both AUC-ROC score and macro-F score are reported. } \label{fig:lambda}
  \setlength{\abovecaptionskip}{0cm}
\end{figure}

In this part, the hyper-parameter $\lambda$ is varied to test {\method}'s sensitivity towards it. To keep simplicity, we adopt ${\method}_T$ and ${\method}_{preT}$ as base model, and set $\lambda$ to be in $\{1e-7,1e-6,2e-6,4e-6,6e-6,8e-6,1e-5\}$. Each experiment is conducted on Cora with imbalance ratio $0.5$ and over-sampling scale $2.0$. The results were shown in Figure~\ref{fig:lambda}. From the figure, we can observe that: (i) Generally, as $\lambda$ increases, the performance first increase then decrease. The performance would drop significantly if $\lambda$ is too large. Generally, a smaller $\lambda$ between $1e-6$ and $4e-6$ works better. The reason could be the difference in scale of two losses; and (ii) Pre-training makes {\method} more stable w.r.t $\lambda$.
% \begin{itemize}
%     \item Generally, as $\lambda$ increases, the performance first increase then decrease. The performance would drop significantly if $\lambda$ is too large. Generally, a smaller $\lambda$ between $1e-6$ and $4e-6$ works better. The reason could be the difference in scale of two losses.
%     \item Pre-training makes {\method} more stable w.r.t $\lambda$.
% \end{itemize}
% \begin{itemize}
%     \item $\lambda$ should better be set as a small value. The performance would drop significantly if it is too large. The reason could be the difference in scale of two losses.
%     \item After pre-training, {\method} would perform much more stable.
% \end{itemize}

\subsection{Performance of GraphSMOTE with Mixup}~\label{sec:expmixup}
In this subsection, we analyze the performance of {\method} with mixup  introduced in Section~\ref{sec:mixup}. With this extension, nodes of majority classes will also be utilized during the synthesis of new instances, and we conduct a series of experiments to evaluate its contribution.

\subsubsection{Settings}
In this part, we introduce hyper-parameter settings involved in the mixup process. Weight of $\mathcal{L}_{mix}$, $\lambda_2$, is fixed as $0.1$. The mixup ratio, which means the ratio of nodes generated via mixup and can be larger then $1$, is set to $1.0$. The interpolation scale in mixup, $b$, is set as $0.5$. Other configurations are the same as introduced in Section $6.1.4$. Throughout experiments, all settings remain fixed unless stated otherwise.

Basing on whether the optional pseudo label obtention process is conducted, we implement two variants and annotate them as $Mix$ and $Mix'$ respectively. 
\begin{itemize}
    \item $Mix'$ means that no pseudo labels are generated, and only those labeled nodes are selected for graphmix. 
    \item $Mix$ means that nodes with pseudo labels are also selected, and the threshold $\mathcal{T}$ is set as $0.3$.
\end{itemize} 
For baseline comparison, we also implement graph mixup alone without {\method} on the imbalance learning setting, to evaluate its contributions.

\subsubsection{Base Result} 
In this part, we evaluate the performance of mixup in the conventional imbalance setting, without the few-label constraint. We test the advantage of mixup on all the three datasets, with the imbalance ratio of Cora set as $0.5$. BlogCatalog is naturally imbalanced and we does not manipulate it. All experiments are randomly conducted for $3$ times with both mean and standard deviations reported. 

The result are summarized in Table~\ref{tab:mix}. Since an accurate relation estimation is required for inserting mixed nodes into the raw graph, we only incorporate this mixup module under the pretrained setting. $Pre+Mix$ means that after pretraining, only mixup technique is used. $PreT+Mix$ and $PreO+Mix$ refer to apply mixup on ${\method_{preT}}$ and ${\method_{preO}}$ respectively. From the result, we can make an observation that:
\begin{itemize}
    \item Synthesizing mixed nodes shows some benefits compared to only interpolating nodes of the same class. For example, with ${{\method}_{preT}}$ as the base model, using $Mix$ can further increase the accuracy by $1.5\%$ and increase the F score by $1.4\%$ on Cora. 
    \item Obtaining pseudo labels and including them in the mixup process has the potential of improving the performance further. $Mix$ consistently outperforms $Mix'$ w.r.t both accuracy and macro F score. Based on this observation, we always use $+Mix$ in the following experiments.
    \item Comparing $Pre+Mix$ with $Origin$ in Table~\ref{tab:result}, we can find that using graph mixup technique alone is also beneficial for the imbalance node classification task, although not as effective as directly synthesizing nodes of minority classes.
\end{itemize}

Although mixup is shown to be beneficial in this setting, the improvement is relatively marginal. We attribute this behavior to that mixup is less required when supervised nodes are sufficient. Next, we will analyze the advantage of mixup augmentation in the few-label scenarios, and when imbalance ratio is more extreme.

\begin{table*}[t]\scriptsize
  \setlength{\tabcolsep}{4.5pt}
  
  \caption{Evaluation of Mixup technique with imbalance ratio $0.5$ and class size $20$. Although it is shown to be beneficial, improvements are small. Later, we test on few-shot and extremely-imbalanced scenarios, and Mixup achieves clearer improvements in both cases.  }\label{tab:mix} 
  \vskip -1em
  \begin{tabular}{p{2.4cm} | p{1.12cm}  p{1.52cm}  p{1.52cm} | p{1.12cm}  p{1.52cm}  p{1.52cm} | p{1.12cm}  p{1.52cm}  p{1.52cm} }

    \hline
     &  \multicolumn{3}{|c|}{Cora} &  \multicolumn{3}{|c|}{BlogCatalog} &  \multicolumn{3}{|c}{Twitter} \\
    \hline
    Methods & ACC & AUC-ROC & F Score & ACC & AUC-ROC & F Score & ACC & AUC-ROC & F Score \\
    \hline
    $Pre+Mix'$ & $69.7\pm1.0$ & $0.927\pm0.004$ & $0.689\pm0.005$ & $20.4\pm0.4$ & $0.606\pm0.001$ & $0.080\pm0.002$ & $91.7\pm0.5$ & $0.611\pm0.003$ & $0.513\pm0.002$  \\
    $Pre+Mix$ & $70.2\pm0.9$ & $0.931\pm0.005$ & $0.704\pm0.004$ & $20.3\pm0.6$ & $0.604\pm0.004$ & $0.078\pm0.003$ & $92.1\pm0.7$ & $0.614\pm0.004$ & $0.516\pm0.003$ \\
    \hline
    $GSMOTE_T$ & $71.3\pm0.8$ & $0.929\pm0.006$ & $0.720\pm0.002$ & $20.6\pm0.5$ & $0.602\pm0.004$ & $0.083\pm0.003$ & $92.9\pm0.5$ & $0.622\pm0.003$ & $0.519\pm0.001$ \\
    $GSMOTE_O$ & $70.9\pm1.0$ & $0.927\pm0.011$ & $0.712\pm0.003$ & $21.5\pm1.0$ & $0.591\pm0.012$ & $0.080\pm0.005$ & $90.5\pm0.8$ & $0.616\pm0.006$ & $0.515\pm0.003$ \\
    $GSMOTE_{preT}$ & $72.7\pm0.3$ & $0.931\pm0.002$ & $0.726\pm0.001$  & ${24.9}\pm0.2$ & ${0.641}\pm0.001$ & ${0.126}\pm0.001$ & $93.7\pm0.3$ & ${0.639}\pm0.002$ & $0.531\pm0.001$ \\
    $GSMOTE_{preO}$ & ${73.6}\pm0.1$ & ${0.934}\pm0.002$ & ${0.727}\pm0.001$ & $24.3\pm0.2$ & ${0.641}\pm0.002$ & $0.123\pm0.001$ & $94.1\pm0.2$ & $0.636\pm0.001$ & ${0.532}\pm0.001$ \\
    \hline
    ${preT}+Mix'$ & $73.6\pm0.5$ & $0.944\pm0.002$ & $0.733\pm0.002$ & $24.9\pm0.1$ & $0.642\pm0.001$ & $0.127\pm0.001$ & $94.0\pm0.5$ & $0.639\pm0.003$ & $0.532\pm0.002$  \\
    ${preO}+Mix'$ & $73.9\pm0.3$ & $0.944\pm0.003$ & $0.730\pm0.001$ & $24.6\pm0.2$ & $0.640\pm0.001$ & $0.124\pm0.001$ & $94.3\pm0.3$ & $0.638\pm0.002$ & ${0.531}\pm0.003$ \\
    \hline
    ${preT}+Mix$ & $73.8\pm0.4$ & $0.945\pm0.003$ & $\underline{0.736}\pm0.001$  & $\underline{25.1}\pm0.2$ & $0.641\pm0.002$ &  $\underline{0.129}\pm0.001$ & $94.4\pm0.2$ & $0.641\pm0.002$ & ${0.535}\pm0.001$  \\
    ${preO}+Mix$ & $\underline{74.2}\pm0.3$ & $\underline{0.947}\pm0.002$ & $0.732\pm0.002$ & $24.8\pm0.3$ & $\underline{0.644}\pm0.001$ & $0.126\pm0.002$ & $\underline{94.6}\pm0.3$ & $\underline{0.642}\pm0.003$ & $\underline{0.536}\pm0.002$ \\
    \hline
    
  \end{tabular}
\end{table*}

\subsubsection{Influence of Imbalance Ratio}
In this subsection, we analyze the performance of mixup extension with respect to different imbalance ratios to evaluate the improvement, especially in the more extreme imbalanced cases. Experiment is conducted in a well-constrained setting on Cora, by fixing over-sampling scale to $1.0$, mixup ratio to $1.0$, and mixup weight $\lambda_2$ to $0.1$. Imbalance ratio is varied as $\{0.1,0.2,0.4,0.6\}$. Each experiment is conducted 3 times and the average results in terms of AUC are shown in Table~\ref{tab:MixImRatio}. From the table, we observe:
\begin{itemize}
    \item Mixup technique improves node classification performance stably across all imbalance ratios. It brings improvements on both variants of {\method}.
    \item Generally, the contribution of mixup is larger when the class is more imbalanced. Taking ${\method_{preT}}$ as an example, mixup improves the AUC score by $2.7\%$ when imbalance ratio is $0.2$, and by $1.5\%$ when imbalance ratio is $0.6$.
    %\item When the imbalance ratio becomes extreme, say $0.1$, the improvement may also decrease, which is because few minority instances are not representative of the data distribution, hence generated mixed nodes are not smoothly distributed on the real classification boundary.
\end{itemize}

These results validate the effectiveness of mixup extension, especially when the number of labeled nodes are small for minority classes.

\begin{table}[t!]
  \setlength{\tabcolsep}{4.5pt}
  \small
  \caption{Node classification performance in terms of AUC on Cora under various imbalance ratios.} \label{tab:MixImRatio}
  \vskip -1em
  \begin{tabular}{c || c | c | c | c  }
    \hline
     &  \multicolumn{4}{|c}{Imbalance Ratio} \\
    \hline
    Methods & $0.1$ & $0.2$ & $0.4$ & $0.6$ \\
    \hline
    ${\method}_T$ & $0.8824$ & $0.9162$ & $0.9262$ & $0.9309$ \\
    ${\method}_O$ & $0.8849$ & $0.9061$ & $0.9216$ & $0.9311$ \\
    ${\method}_{preT}$ & $0.9167$ & $0.9130$ & $0.9303$ & $0.9317$ \\
    ${\method}_{preO}$ & $0.9117$ & $0.9116$  & $0.9389$ & $0.9366$ \\
    \hline
    $preT+Mix$ & $0.9206$& $0.9378$ & $0.9434$ & $0.9458$ \\
    $preO+Mix$ & $\mathbf{0.9293}$ & $\mathbf{0.9412}$ & $\mathbf{0.9451}$ & $\mathbf{0.9512}$ \\
    \hline
    
  \end{tabular}
\end{table}

\subsubsection{Few-labeled Scenario}
\begin{figure}[t!]
  \centering
  \subfigure[$preT+Mix$]{
		
		\includegraphics[width=0.23\textwidth]{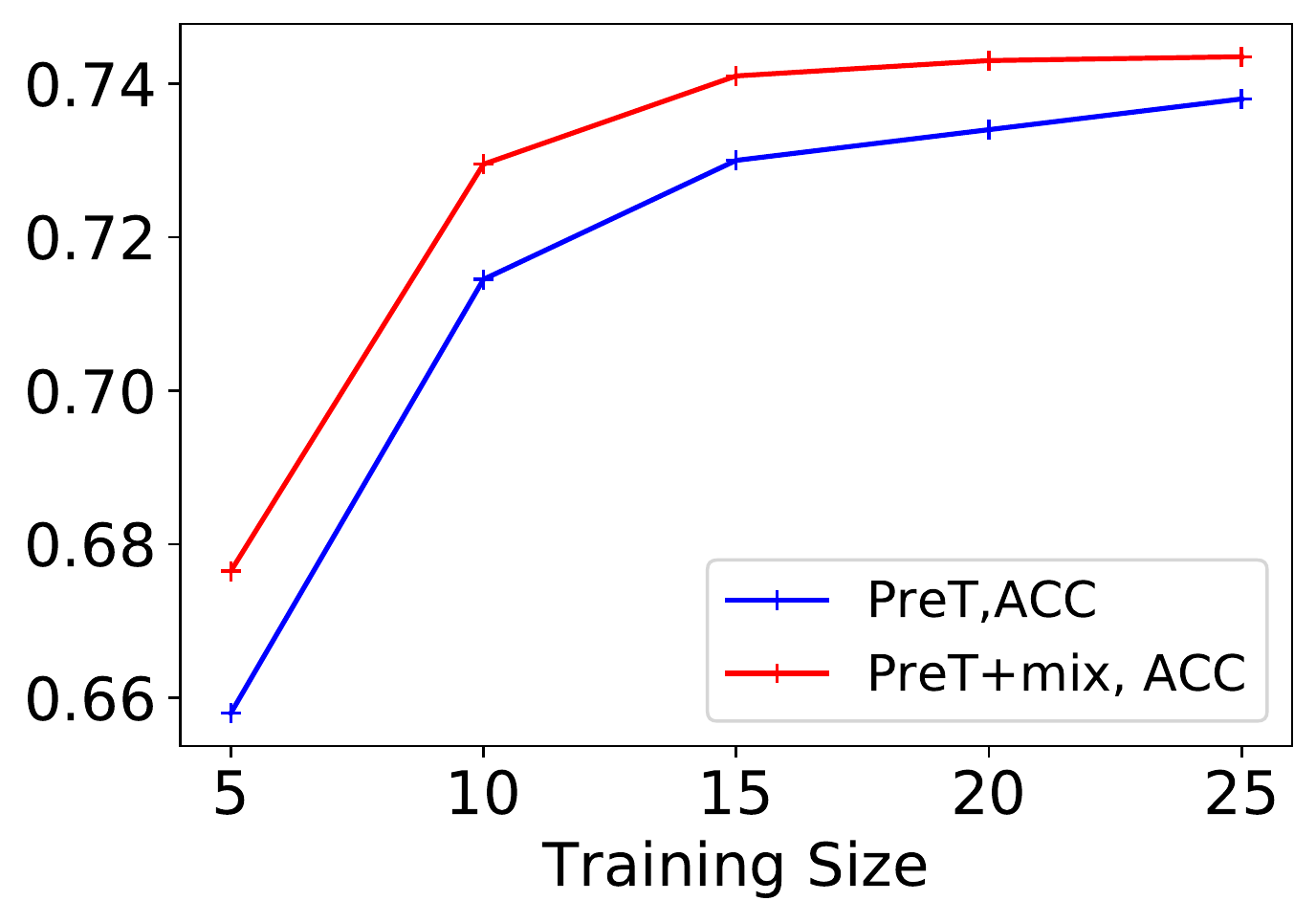}}
    \subfigure[$preO+Mix$]{
		\includegraphics[width=0.23\textwidth]{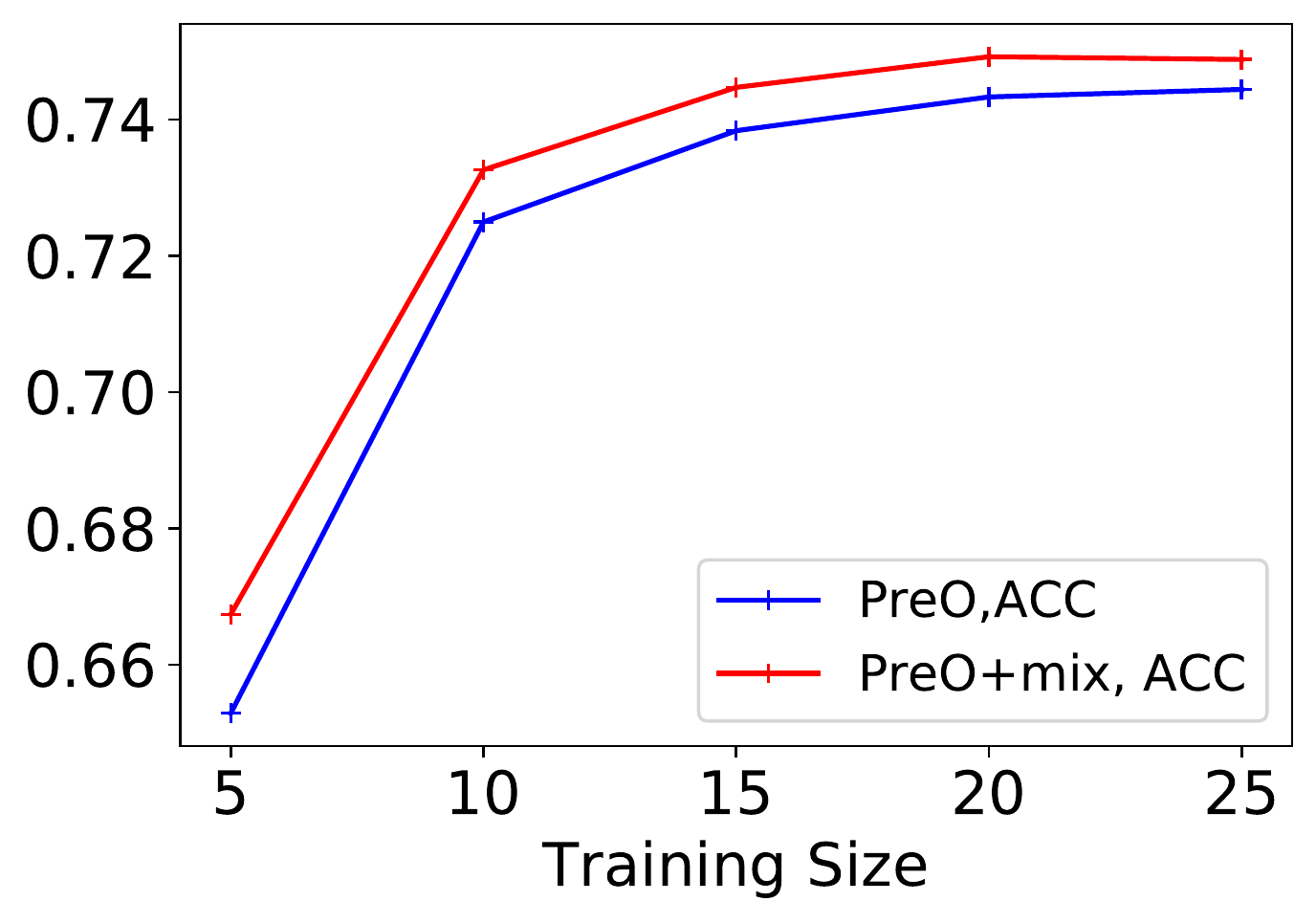}}
		
    \vskip -1em
    \caption{Analysis of the mixup extension on {\method} with respect to different amount of labeled training data.}\label{fig:size}
  \setlength{\abovecaptionskip}{0cm}
\end{figure}

Through mixup, we can utilize both positive and negative nodes during augmenting the given graph, which is important when we have only limited number of labeled nodes. In this experiment, we keep the imbalance ratio as $0.5$, and vary the number of labeled nodes in majority class as $\{5, 10, 15, 20, 25\}$ to analyze its contribution in the few-label scenario. Mixup ratio is fixed as $3.0$, . We leave other configurations like interpolation scale, mixup loss weight, and $\mathcal{T}$ unchanged. Experiments are conducted on Cora dataset for three times, and we show the result in terms of accuracy in Figure~\ref{fig:size}. From the figure, we can see that mixup is more effective in the few-labeled case, and when the amount of labeled nodes is rich, its improvement is smaller. 

This observation matches expectations. Mixup technique is introduced to address the semi-supervision scenario, in which only a small ratio of nodes are labeled and available during training. With supervision ratio being small, using majority nodes to improve the data augmentation process would be more important. When supervision ratio becomes larger, this design is no longer as necessary.

\subsubsection{Influence of Mixup Ratio}
\begin{figure}[t!]
  \centering
  \subfigure[$preT+Mix$]{
		
		\includegraphics[width=0.23\textwidth]{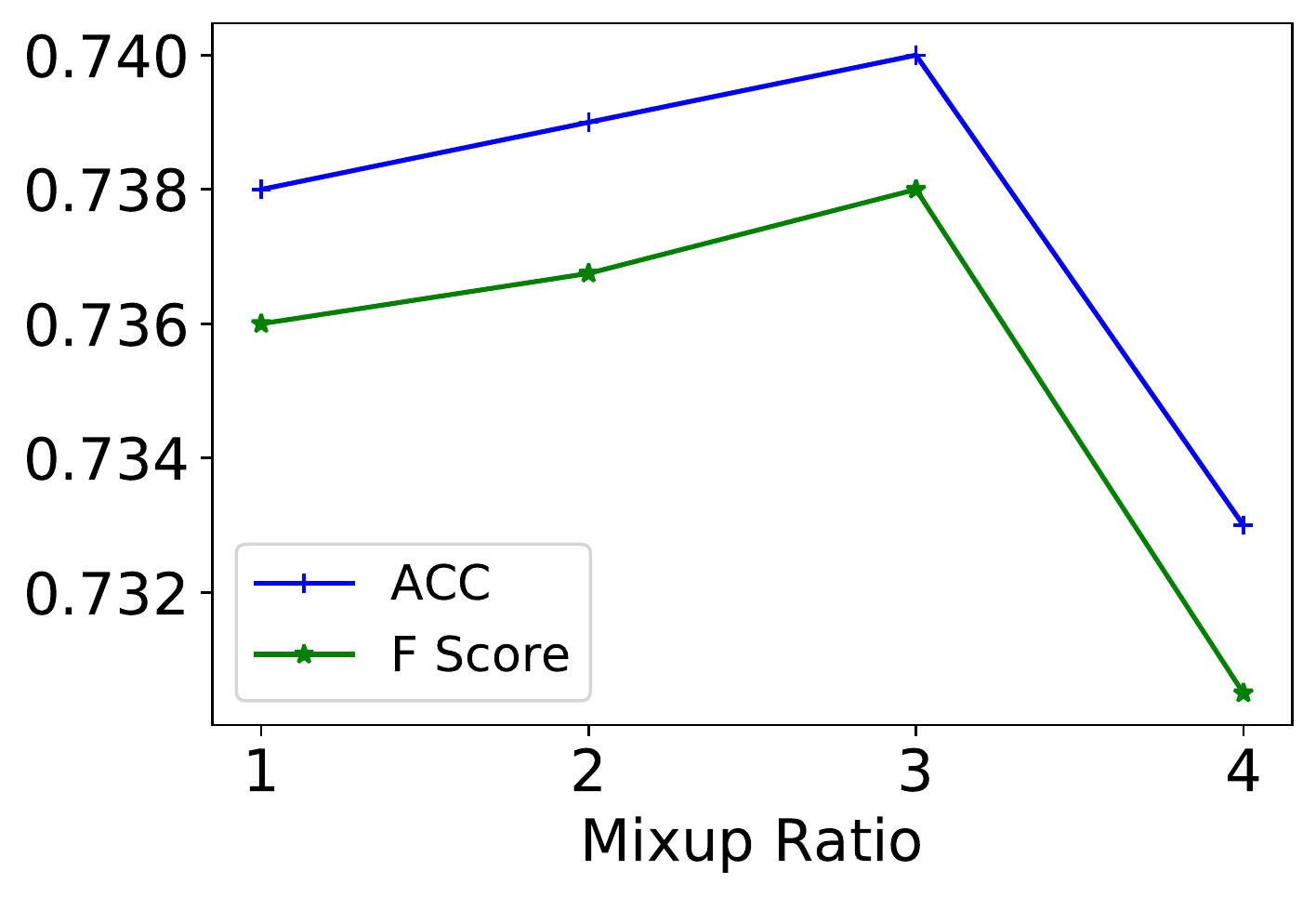}}
    \subfigure[$preO+Mix$]{
		\includegraphics[width=0.23\textwidth]{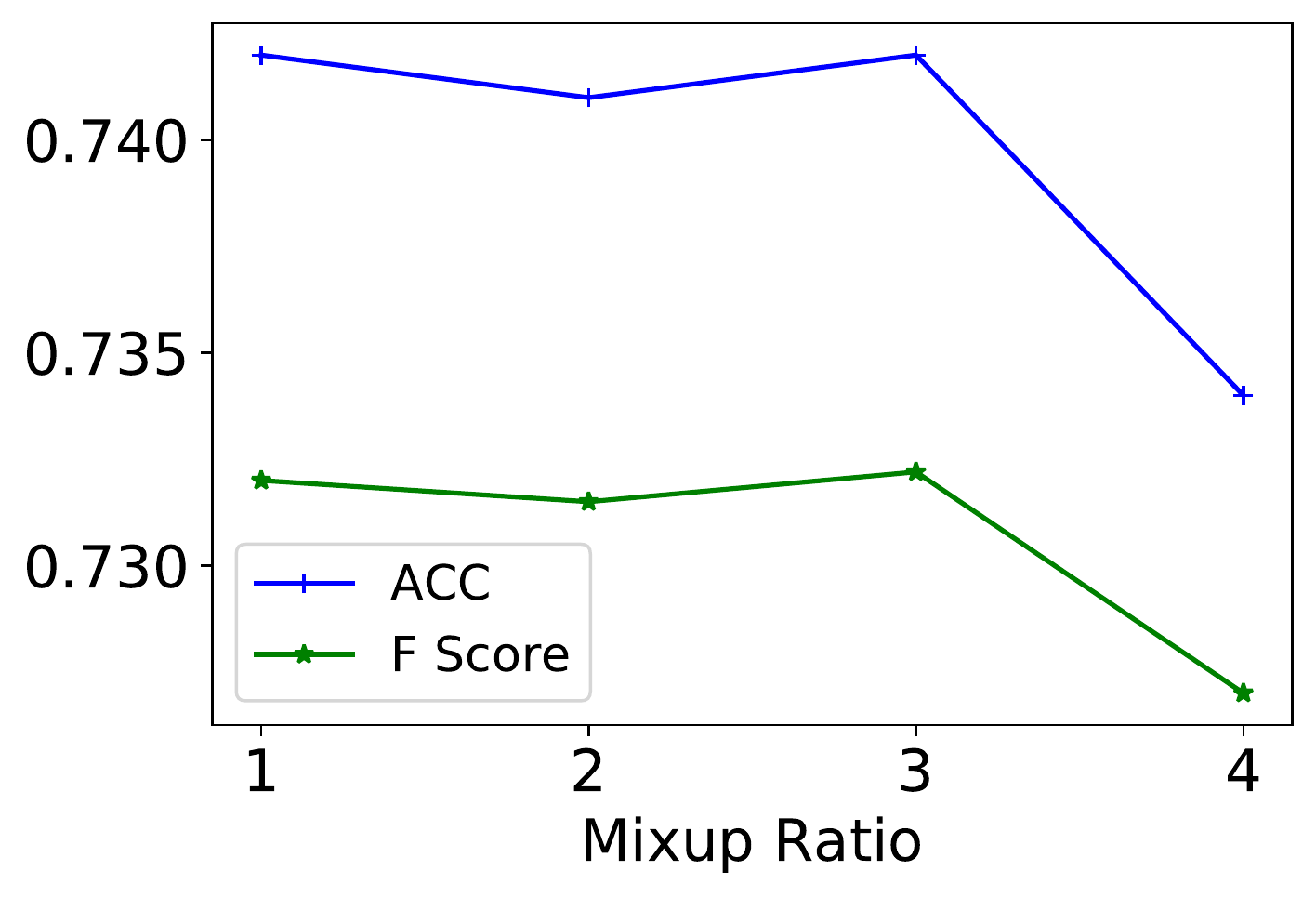}}
		
    \vskip -1em
    \caption{Sensitivity analysis of mixup technique on mixup ratio. Higher ratio means that more mixed nodes are generated and inserted into the graph, and results on both mean accuracy score and macro-F score are reported.}
  \setlength{\abovecaptionskip}{0cm}\label{fig:mixratio}
\end{figure}

In this subsection, we evaluate the sensitivity of the performance w.r.t the amount of generated mixed nodes, measured by mixup ratio. A larger mixup ratio would provide more augmented nodes for training, but may also introduce larger noises. To conduct experiments in a constrained setting, we use Cora dataset, and keep all other configurations the same as main experiment. Mixup ratio is varied as $\{1.0, 2.0, 3.0. 4.0\}$,  and every experiment is randomly conducted 3 times. Average resultsin terms of accuracy and F score are presented in Figure~\ref{fig:mixratio}. From the result, two observations can be made:
\begin{itemize}
    \item It is recommended to set mixup ratio between $[1.0,3.0]$. Increasing mixup ratio to be larger than $3.0$ will have a negative influence on the performance;
    \item Compared to using ${\method_{preT}}$ as the base model, adding mixup technique to ${\method_{preO}}$ is less sensitive towards mixup ratio. We attribute this to the fact that ${\method_{preT}}$ directly utilizes generated mixed graph while ${\method_{preO}}$ can learn to dynamically update it. As a result, the performance of ${\method_{preT}}$ is more dependent on the augmented graph than  ${\method_{preO}}$.
\end{itemize}

\subsubsection{Influence of Interpolation Scale}
\begin{figure}[t!]
  \centering
  \subfigure[$preT+Mix$]{
		
		\includegraphics[width=0.23\textwidth]{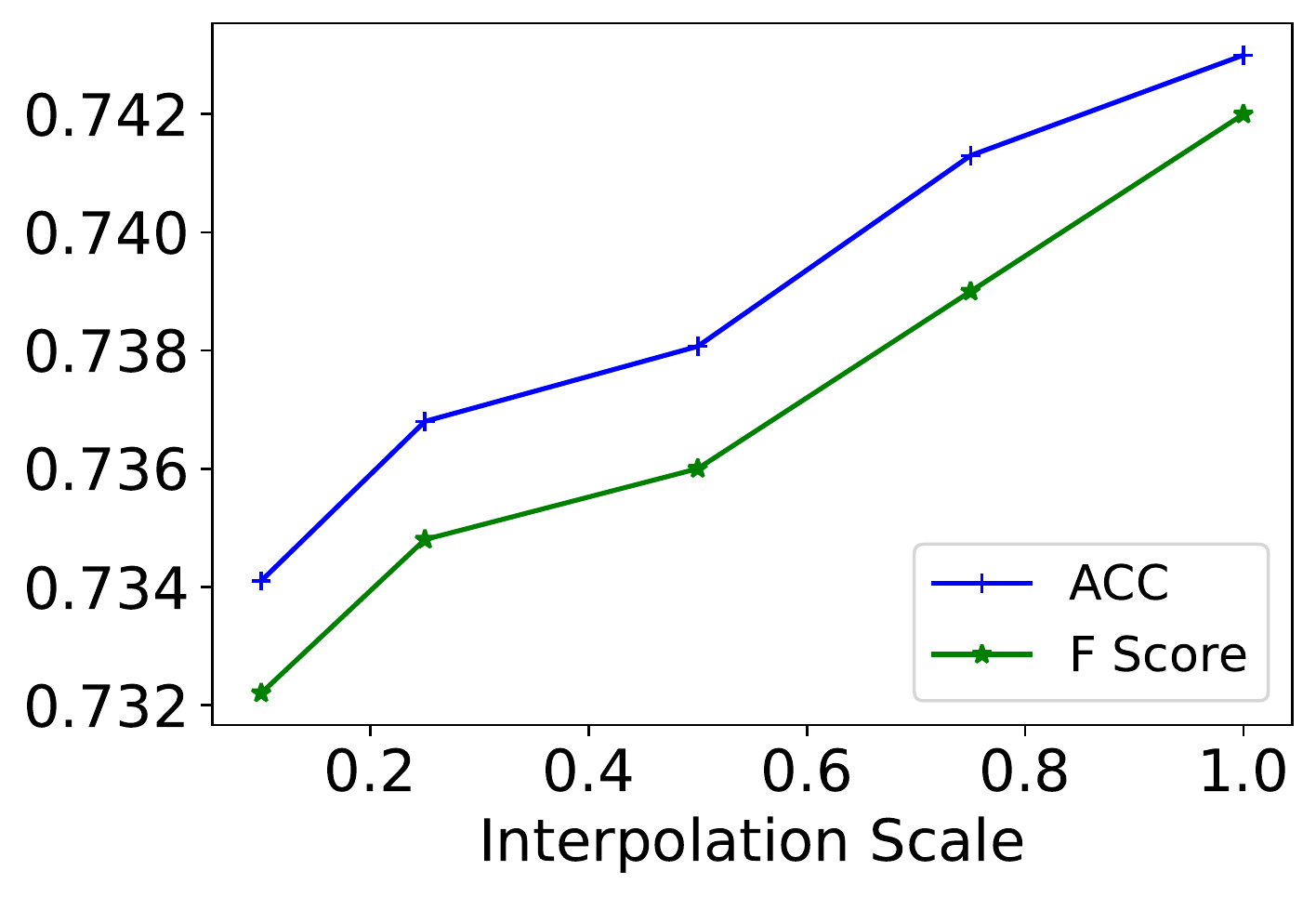}}
    \subfigure[$preO+Mix$]{
		\includegraphics[width=0.23\textwidth]{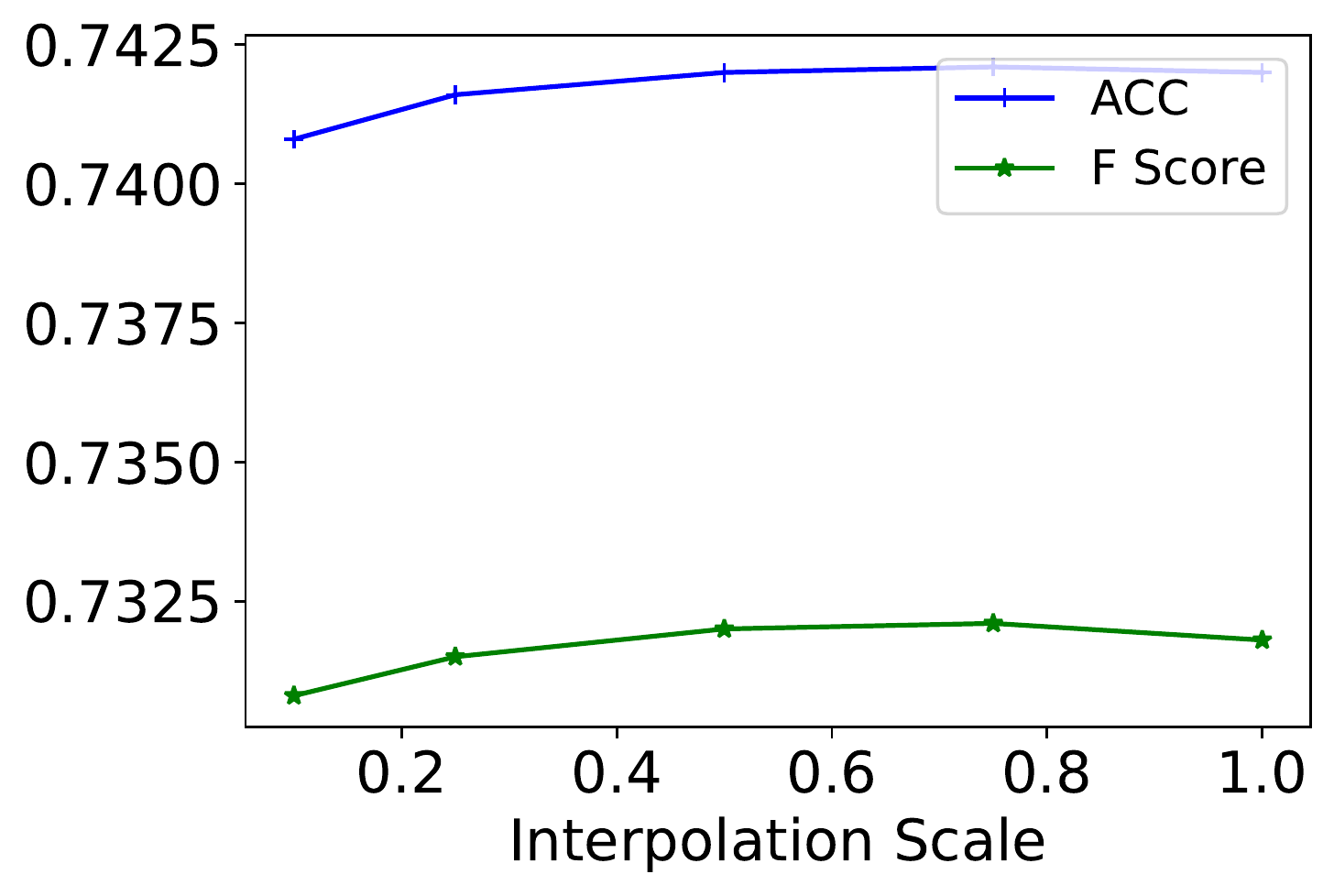}}
		
    \vskip -1em
    \caption{Sensitivity analysis of mixup technique on interpolation scale. Larger scale means that different node classes are mixed more evenly, and results on both mean accuracy score and macro-F score are reported.}\label{fig:interp}
  \setlength{\abovecaptionskip}{0cm}
\end{figure}

In this subsection, we evaluate the sensitivity of the performance w.r.t the interpolation scale $b$, which influences the distribution of generated new nodes. A larger scale will make different classes mixed more evenly. Cora dataset is adopted for the experiment, and all other configurations remain the same as previous experiments. Interpolation scale is varied as $\{0.1, 0.25, 0.5, 0.75, 1.0\}$, and every experiment is randomly conducted 3 times. Average results in accuracy and F score are presented in Figure ~\ref{fig:interp}. From the figure, it is shown that increasing interpolation scale and generating more ``in-between'' nodes are beneficial for the performance of proposed method. Besides, the influence is more distinct on ${\method_{preT}}$ compared to ${\method_{preO}}$. 

\subsubsection{Influence of Mixup Loss Weight}
\begin{figure}[t!]
  \centering
  \subfigure[$preT+Mix$]{
		
		\includegraphics[width=0.23\textwidth]{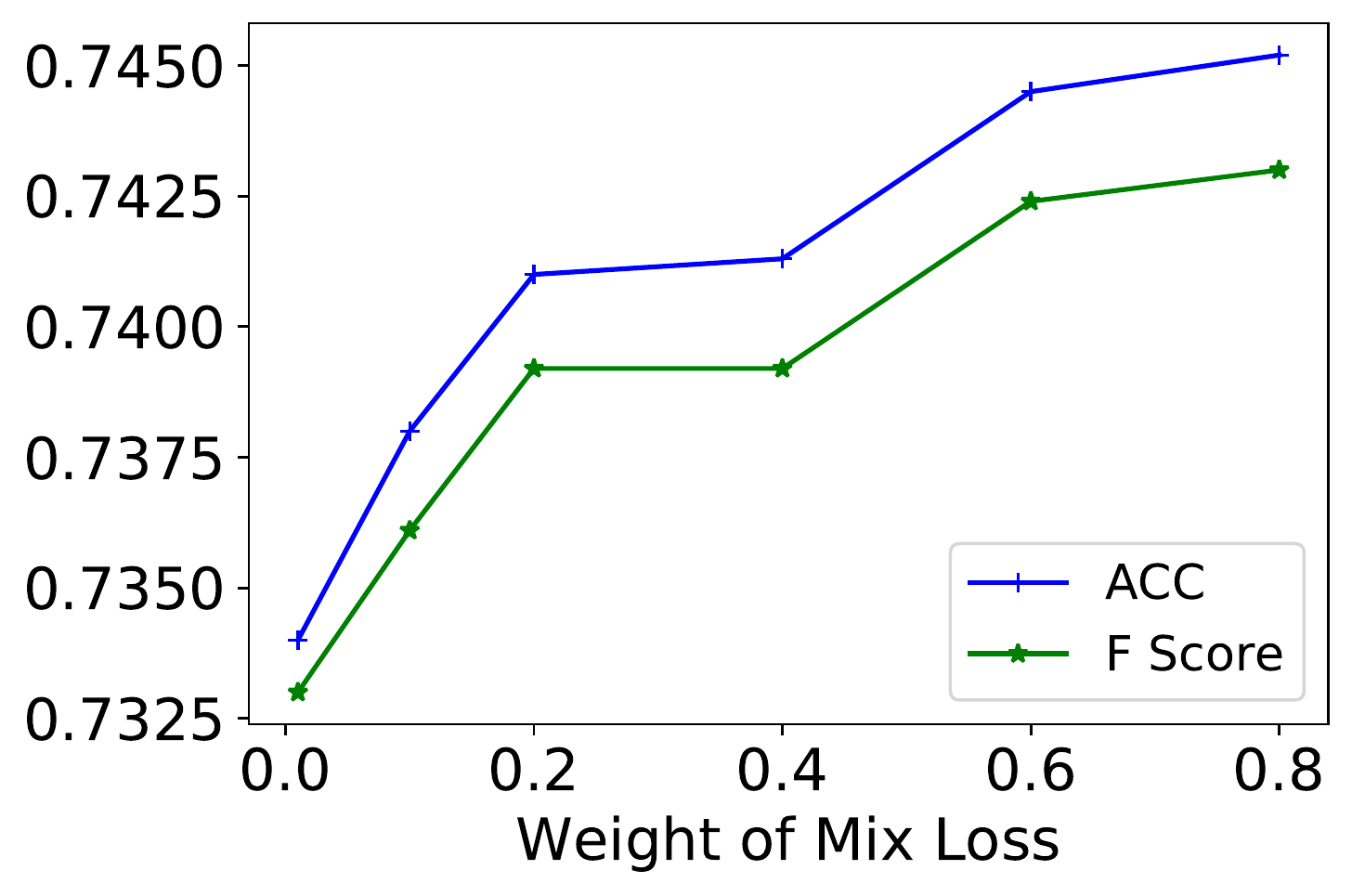}}
    \subfigure[$preO+Mix$]{
		\includegraphics[width=0.23\textwidth]{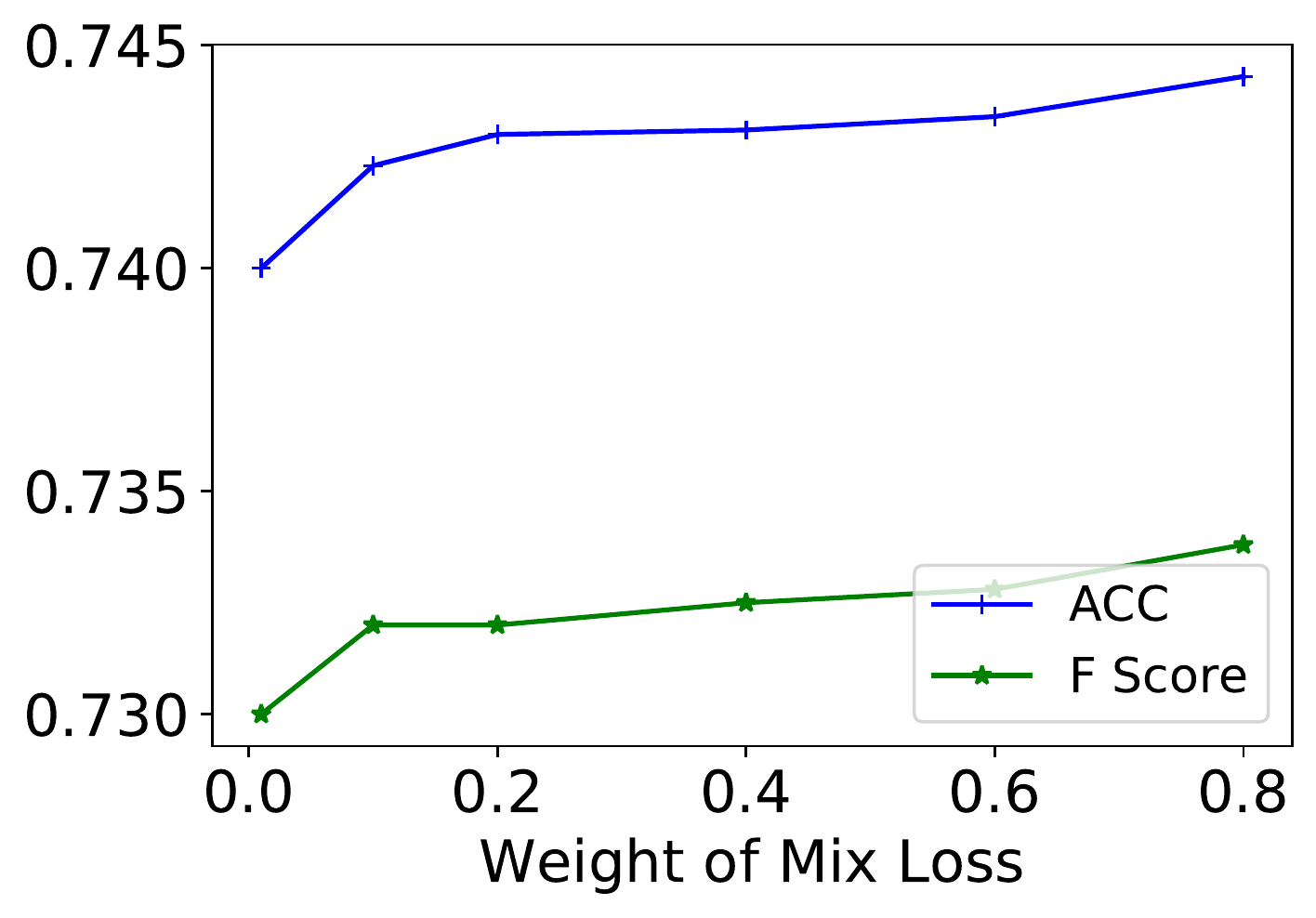}}
		
    \vskip -1em
    \caption{Ablation study of mixup technique on {\method} framework, through changing the weight of mixup loss, $\lambda_2$. Performance is evaluated using both mean accuracy score and macro-F score.}\label{fig:weight}
  \setlength{\abovecaptionskip}{0cm}
\end{figure}

In this subsection, to analyze the balance between classification on supervised nodes and on generated mixed nodes, we test the sensitivity of model's performance w.r.t the weight of $\mathcal{L}_{mix}$, $\lambda_2$. Experiments are conducted on Cora dataset, and all other configurations remain unchanged. $\lambda_2$ is varied as $\{0, 0.01, 0.1, 0.2, 0.4, 0.6, 0.8\}$, and every experiment is randomly conducted 3 times. Average results in both accuracy and F score are presented in Figure: ~\ref{fig:weight}. From the result, we can observe that the influence of increasing mixup loss weight is clearer within the range $[0.0, 0.2]$, and the curve would become smoother when $\lambda_2$ goes up further. Besides, in agreement with observations from previous experiments, it can be seen that  ${\method_{preO}}+Mix$ is less sensitive towards changes of hyper-parameters. 

\subsubsection{Influence of Predicted Edges}
In this subsection, we further test the importance of providing relation information for generated mixed nodes. We implement three node insertion approaches:
\begin{itemize}
    \item Vanilla Mix. After generating new mixed nodes following Equation~\ref{eq:mix}, we take them as independent with existing nodes and directly add them to the graph.
    \item Heuristic Mix. In this baseline, we do not use a trained edge predictor to generate relation information for mixed nodes. Instead, we take a heuristic approach: using mixed edges: $\tilde{\mathbf{A}}[\hat{v},:] = (1-\delta') \cdot \mathbf{A}[v,:] + \delta' \cdot \mathbf{A}[u,:]$.
    \item Mix via Prediction. The proposed one, which utilizes a trained edge predictor to provide relation information for synthesized node set $\hat{\mathcal{V}}$.
\end{itemize}

We apply these three approaches with both ${{\method}}_{preT}$ and ${{\method}}_{preO}$ as the base model respectively, and experiments are conducted for $3$ times. Based on observations from previous experiments, mixup ratio is set as $3.0$, and other configurations are remain unchanged. Results are summarized in Table ~\ref{tab:mix_edge}. From the results, we can draw two conclusions:
\begin{itemize}
    \item Inserting mixed nodes into the graph via an auxiliary edge predictor is effective, and it may introduce less noises. For example, with ${\method}_{preO}$ as the base model, using predicted edges shows a clear improvement on both accuracy and macro-F score compared to using heuristic mixed edges.
    \item Although providing relation information for generated mixed nodes is beneficial, the improvement is not that significant most of the time when used in together with ${\method}$ and pre-training. 
\end{itemize}

\begin{table}[t!]
  \setlength{\tabcolsep}{4.5pt}
  
  \caption{Evaluation of node insertion techniques. } \label{tab:mix_edge}
  \vskip -1em
  \begin{tabular}{c || c | c | c   }
    \hline
     &  \multicolumn{3}{|c}{Cora} \\
    \hline
    Methods & ACC & AUC-ROC & F Score  \\
    \hline
    $preT+Vanilla$ & $73.7\pm0.4$ & $0.938\pm0.003$ & $0.735\pm0.003$ \\
    $preT+Heuristic$ & $73.8\pm0.3$ & $0.940\pm0.002$ & $0.735\pm0.004$ \\
    $preT+Pred$ & $74.0\pm0.3$ & $0.942\pm0.002$ & $\mathbf{0.738}\pm0.004$   \\
    \hline
    $preO+Vanilla$ & $74.0\pm0.3$ & $0.941\pm0.002$ & $0.730\pm0.004$ \\
    $preO+Heuristic$ & $73.3\pm0.4$ & $0.939\pm0.002$ & $0.722\pm0.003$  \\
    $preO+Pred$ & $\mathbf{74.2}\pm0.3$ & $\mathbf{0.948}\pm0.001$ & $0.732\pm0.002$   \\
    
    \hline
  \end{tabular}
\end{table}